\documentclass[11pt,letterpaper,logo,onecolumn,nocopyright]{microsoft}

\usepackage[authoryear,round]{natbib}
\usepackage{nicefrac}
\usepackage{algorithm}
\usepackage{algorithmic}
\usepackage{subcaption}
\usepackage{tikz}
\usetikzlibrary{calc}
\usepackage{pgfplots}
\pgfplotsset{compat=1.18}
\usepackage{multirow}

\newtheorem{theorem}{Theorem}

\newtheorem{proposition}[theorem]{Proposition}
\newtheorem{corollary}[theorem]{Corollary}

\newcommand{\cX}{\mathcal{X}}

\newcommand{\cC}{\mathcal{C}}
\newcommand{\cF}{\mathcal{F}}

\newcommand{\eps}{\varepsilon}

\title{Elicited Adaptation: Auditable Localized Fairness via Pairwise Queries}
\hypersetup{pdftitle={Elicited Adaptation: Auditable Localized Fairness via Pairwise Queries}}

\author{Shrey Shah}
\hypersetup{pdfauthor={Shrey Shah}}

\begin{document}
\begin{abstract}
How much must a predictor's scores change when a community's elicited fairness standard changes? Similarity and tolerance judgments define protected comparisons and acceptable score gaps. We study minimum uniform score movement on a fixed pool of $n$ profiles. For scores satisfying a source standard, we characterize the exact cost of retaining that standard during adaptation: runs of consecutive target constraints along shortest paths determine the optimal amplification factor. Its sharp worst case is $\lfloor n/2\rfloor$, and a full cost curve describes finite target-only repair budgets. Committing before the target is known can cost $n-1$ times the worst repair after revelation, even within one fixed convex ball of two-dimensional Mahalanobis metrics. Positive edge-budget floors cap both factors; bounded geometry also limits the commitment cost. At two interior targets of the construction attaining the $n-1$ bound, similarity queries alone yield a feasible $2$-approximate repair for tolerance $\eps\le1/[2(n-1)]$, reducing movement relative to common repair by a factor linear in $n$. Feasible $(1+\alpha)$-optimal repair on this family has worst-case tolerance-query complexity $\Theta(\alpha^{-2}n^2\log(1/\delta))$ under a fixed-slope logistic model. The analysis separates the information needed for a substantial adaptation benefit from the precision needed for nearly optimal movement. Elicitation studies and training comparisons examine tolerance choice and the relation between numerical compliance and predictive utility.
\end{abstract}

\maketitle
\section{Introduction}
\label{sec:intro}

Individual fairness requires similar treatment of similarly situated people~\citep{dwork2012fairness}. Communities may disagree about which comparisons deserve protection and how different the corresponding scores may be. These choices can also change as a system moves to a new setting. Adapting a predictor then requires answering a quantitative question: how much must its scores change, and which earlier obligations can the updated scores retain?

Elicitation introduces a second consideration. While the target standard is uncertain, an updated score vector may need to satisfy every standard consistent with the available judgments. Further queries can rule out candidate targets and reduce this commitment cost. Source obligations, when retained, remain binding throughout adaptation. We analyze how these two requirements affect the necessary score movement.

We minimize the largest score change on a fixed pool of $n$ profiles. Similarity judgments determine protected pairs; tolerance judgments assign their acceptable score gaps. Together they define a graph of obligations. Classical distance envelopes compute an optimal repair for a given graph~\citep{mcshane1934extension,mustata2007}. This yields a constructive post-processing rule and a lower bound on the movement of any fine-tuned model satisfying the same pool constraints. We use this geometry to characterize the additional costs of source retention and unresolved targets, and the precision with which a standard must be elicited.

\paragraph{Exact adaptation cost curve.}
Fix a source graph and a finite family of target graphs. Among source-feasible predictors for which each target can separately be satisfied within movement $a$, we characterize the largest movement needed to satisfy the source and all targets at once. The characterization holds for every $a$ (Theorem~\ref{thm:joint_cost_curve}). A shortest-path closure computes this cost curve and produces predictors attaining it. Its initial slope is an explicit path statistic: the minimum number of target-standard runs along shortest union paths, maximized over relevant endpoint pairs. Source segments contribute zero to this count. This coefficient is the smallest one valid for the fixed family, and it is attained arbitrarily close to target feasibility. The full curve is continuous and piecewise linear; it can alternate between flat and increasing segments, so its initial amplification factor alone need not describe finite-budget adaptation.

\paragraph{Sharp source-retention and commitment bounds.}
With one target, the path statistic reduces to the number of target runs, giving the sharp worst-case factor $\lfloor n/2\rfloor$. With multiple unresolved targets, the sharp factor is $n-1$. These are special cases of the same characterization. Both factors have realizations using two-dimensional Mahalanobis metrics. For the commitment factor, one fixed convex ball of positive-definite matrices works for every pool size: individual targets give alternating matchings, while their common constraints form a path. Positive lower bounds on edge budgets cap the factors independently of $n$; bounded profile geometry also caps the commitment factor under a complete pair domain and a common tolerance. The constructions show how alternating constraints increase adaptation costs, while the bounds identify conditions that limit this increase.

\paragraph{Tolerance-query complexity and repair geometry.}
Confidence bounds on unresolved constraints can certify a movement budget before the whole standard is identified. Once the target graph is known, we characterize how its repair radius changes with a common tolerance, including when source obligations remain binding. Critical paths determine sensitivity to tolerance error and yield response bounds for a feasible $(1+\alpha)$-optimal repair under a fixed-slope logistic acceptance model. On affine portions of the repair curve, local testing lower bounds have the same dependence on critical target-edge count and the required movement precision. For the alternating-matching family, similarity queries at two interior targets identify the graph using a number of responses independent of $n$, provided the response classes are known. Equalizing the active pairs then gives a feasible $2$-approximation without tolerance responses, whereas nearly optimal movement has worst-case tolerance-query cost $\Theta(\alpha^{-2}n^2\log(1/\delta))$ at fixed noise in score-gap units.

Elicitation studies on COMPAS examine similarity judgments and acceptable score gaps. Training and score-compression comparisons show how tolerance choice changes numerical compliance and predictive utility (Section~\ref{sec:experiments}). Throughout, the selected representation, panel, and pair domain specify the obligations whose adaptation costs we study.

\section{Community-specific fairness specifications}
\label{sec:localized}

Let $h:\cX\to[0,1]$ be a shared score predictor and let $\cC_1,\ldots,\cC_{K_{\rm comm}}$ be a declared partition. Community $k$ has a representation $\phi_k$, a distance $d_k$, and tolerance $\eps_k\ge0$. For Mahalanobis specifications,
\[
 d_k(x,x')=\sqrt{(\phi_k(x)-\phi_k(x'))^\top M_k(\phi_k(x)-\phi_k(x'))},\qquad M_k\succeq0.
\]
The representation is part of the standard: applying the same matrix to different coordinates changes its meaning. A panel supplies judgments about the subjects in a community. Panel composition, community membership, and predictor inputs are distinct design choices.

For a declared domain of within-community pairs, threshold individual fairness requires
\begin{equation}\label{eq:if}
 d_k(x,x')\le1\quad\Longrightarrow\quad |h(x)-h(x')|\le\eps_k.
\end{equation}
A Lipschitz specification instead requires $|h(x)-h(x')|\le\eps_kd_k(x,x')$ on its declared pair domain. It implies~\eqref{eq:if} on pairs with $d_k\le1$ when both domains include those pairs. Linear scaling with distance is an additional modeling assumption beyond an elicited absolute gap.

The partition and pair domain determine which comparisons receive protection. Within-community constraints alone allow a constant score of zero in one community and one in another. Cross-community or demographic-parity guarantees therefore require additional comparisons and assumptions. Combining community rules or eliciting a cross-community rule is a further normative choice. Throughout, compliance is defined relative to the selected partition, representation, panel, and pair domain.

\subsection{Optimal repair and directed adaptation}
\label{sec:transfer_v10}

We formulate adaptation between elicited standards as a graph repair
problem. Fix a pool $V=\{x_1,\ldots,x_n\}$ and existing scores
$h_i=h(x_i)\in[0,1]$. A standard defines a graph
$G=(V,E)$ with finite nonnegative edge budgets $c_{uv}$ and feasible
class
\[
 \cF_G=\{z\in[0,1]^n:|z_u-z_v|\le c_{uv},\ (u,v)\in E\}.
\]
Threshold fairness includes the declared pairs with distance at most one
and uses their elicited tolerances as budgets. A Lipschitz specification
instead uses its declared pair domain and budgets
$c_{uv}=\eps_{uv}d(x_u,x_v)$. Thus both similarity and tolerance enter
the adaptation problem.

\paragraph{Optimal repair of a fixed predictor.}
Let $d_G(u,v)$ be shortest-path distance using the edge budgets as
lengths, infinite across components, and set
$\rho_G(u,v)=\min\{1,d_G(u,v)\}$. Paths encode the implications of local
constraints: $z\in\cF_G$ exactly when
$|z_u-z_v|\le\rho_G(u,v)$ for every pair. Consequently, the minimum
uniform movement required to repair the scores is
\begin{equation}\label{eq:repair_main}
 r_G(h)=\min_{z\in\cF_G}\|z-h\|_\infty
 =\frac12\max_{u,v}[|h_u-h_v|-\rho_G(u,v)]_+.
\end{equation}
Lower and upper distance envelopes give a constructive optimal repair:
\begin{equation}\label{eq:midpoint_main_v6}
 L_u=\min_v\{h_v+\rho_G(u,v)\},\quad
 U_u=\max_v\{h_v-\rho_G(u,v)\},\quad
 z_u^{\rm mid}=(L_u+U_u)/2.
\end{equation}
The midpoint preserves every already-feasible component and is
nonexpansive in $h$ for a fixed graph. Two multi-source Dijkstra searches
compute it in $O((|E|+n)\log n)$ time and $O(|E|+n)$ storage for $n\ge2$,
after graph construction. This construction uses Lipschitz extension
and shortest-path methods~\citep{mcshane1934extension,mustata2007};
Proposition~\ref{prop:exact_pool_repair_v3} and
Appendix~\ref{app:repair_selection_v6} give proofs and alternative
selections among optimal repairs.

\paragraph{Directed comparison of standards.}
The maximum repair cost over source-compliant predictors defines a
directed comparison of the standards:
\begin{align}
 D(s\!\to\!t)
 &:=\sup_{h\in\cF_s}\inf_{z\in\cF_t}\|h-z\|_\infty
 \nonumber\\
 &=\frac12\max_{u,v}[\rho_s(u,v)-\rho_t(u,v)]_+.
 \label{eq:directed_transfer_main_v10}
\end{align}
Scores given by capped source path distances from one vertex attain this
value (Corollary~\ref{cor:graph_transfer_v3}). This quantity depends on
implied path constraints and is invariant to changes in metric parameters
that preserve the feasible class. For two profiles with source budget
$0.4$ and target budget $0.1$, $D(s\!\to\!t)=0.15$, but
$D(t\!\to\!s)=0$. Scores $(0,0.4)$ attain the forward cost through
repair $(0.15,0.25)$; equal source scores require no repair.

\section{Joint analysis of source retention and target uncertainty}
\label{sec:joint_adaptation}

Adaptation may require continued source compliance while the target
standard is still being elicited. We analyze the joint cost of these
requirements. Fix a source graph
$G_s$ and a finite family of target graphs $G_1,\ldots,G_p$, $p\ge1$, on
the same pool. Let $H$ be their union, with the smaller budget on overlaps,
so $\cF_H=\cF_s\cap\bigcap_{j=1}^p\cF_j$. For $h\in\cF_s$, define
\begin{equation}\label{eq:joint_radii}
 R(h)=\max_{1\le j\le p}r_j(h),\qquad Q(h)=r_H(h).
\end{equation}
The revealed cost $R$ allows a separate repair for each target; $Q$
requires a common repair satisfying every target and retaining the source.
Their comparison measures the joint cost of source retention and target
uncertainty.

For a particular predictor, $Q(h)$ is computed directly from the union
graph. The coefficient $K$ and curve $C(a)$ below depend only on the
standards. They give uniform movement guarantees for source-feasible
predictors before one is selected. In particular, $C(a)\le b$ certifies that
every such predictor with target-specific repair costs at most $a$
admits a common repair within deployment budget $b$.

\paragraph{Four-profile example.}
Take scores $(0,1/3,2/3,1)$ and a source path with budget $1/3$ on
each adjacent pair. Target 1 protects $(1,2)$ and $(3,4)$ with budget
$1/6$; target 2 protects $(2,3)$ with the same budget. Either target
alone costs $1/12$ to repair. For target 1, moving each protected pair
inward by $1/12$ opens the middle gap to $1/2$, violating the source.
Retaining the source raises its repair cost to $1/6$, attained by
$(1/6,1/3,2/3,5/6)$. Satisfying both targets restricts the total endpoint
gap to $1/2$ and costs $1/4$, attained by
$(1/4,5/12,7/12,3/4)$. Thus target-specific adaptation, source-retaining
adaptation, and commitment to both targets have distinct costs on the
same four profiles. The path statistic defined below characterizes the
factors of two and three.

\paragraph{Constraint paths and movement amplification.}
Assign each edge of $H$ to a standard attaining its union budget.
Source-assigned edges are already satisfied. Consecutive edges assigned
to the same target contribute at most $2R(h)$ to the endpoint deficit,
independently of the run's length. A change of target begins a new run;
source edges separate runs without contributing a deficit.
For connected $u,v$, let $k(u,v)$ be the minimum number of target runs
over simple shortest $H$-paths and all compatible assignments. Define
\begin{equation}\label{eq:joint_factor}
 K=\max_{d_H(u,v)<1}k(u,v),\qquad k(u,u)=0.
\end{equation}
Only paths with total budget below one can create a positive deficit.
If several standards attain an edge's minimum budget, the assignment may
use any of them.

\begin{theorem}[Exact joint amplification]\label{thm:joint_amplification}
For every source-feasible predictor,
\[
 R(h)\le Q(h)\le\min\{1/2,K R(h)\}.
\]
The coefficient $K$ is the smallest nonnegative one valid for all
$h\in\cF_s$. If $K>0$, every sufficiently small $a>0$ admits a
source-feasible predictor with $R(h)=a$ and $Q(h)=Ka$.
If $K=0$, then $\cF_s\subseteq\bigcap_{j=1}^p\cF_j$.
Always $K\le n-1$; with one target, $K\le\lfloor n/2\rfloor$.
Both bounds are sharp.
\end{theorem}

Summing endpoint deficits once per target run proves the upper bound.
The exact cost curve below proves attainment for each fixed family,
including tied budgets and zero-length edges
(Appendix~\ref{app:joint_adaptation}). A single path statistic therefore
characterizes amplification under joint source retention and target
uncertainty. The following sections establish sharp bounds for two
extremal cases: one known target and two alternating possible targets with
an unconstrained source.

\paragraph{Predictors with residual source violations.}
The comparison also gives a repair budget when training only approximately
enforces the source. For arbitrary $h\in[0,1]^n$, put $s=r_s(h)$ and
$R(h)=\max_jr_j(h)$. Repairing the source first and then applying the
joint comparison gives
\begin{equation}\label{eq:joint_inexact_source}
 r_H(h)\le\min\{1/2,(K+1)s+K R(h)\}.
\end{equation}
Each target repair radius is $1$-Lipschitz in the input scores, so the
first movement of size $s$ increases $R$ by at most $s$.
This separates measured source violations from amplification during
adaptation. When $K=0$, $\cF_H=\cF_s$ and $r_H(h)=s$ exactly.

\paragraph{Exact cost curve.}
Let $S=\rho_s$ and $U(u,v)=\min_{1\le j\le p}\rho_j(u,v)$.
For $a\ge0$, let $\rho_a$ be the shortest-path closure of the complete-pair
budgets $w_a(u,v)=\min\{S(u,v),U(u,v)+2a\}$.
By~\eqref{eq:repair_main}, these constraints describe exactly the
source-feasible predictors with revealed cost at most $a$.

\begin{theorem}[Exact joint cost curve]\label{thm:joint_cost_curve}
The smallest common movement budget valid for all such predictors is
\begin{equation}\label{eq:joint_cost_curve}
 C(a):=\sup_{h\in\cF_s:\,R(h)\le a}Q(h)
 =\frac12\max_{u,v}\{\rho_a(u,v)-\rho_H(u,v)\}.
\end{equation}
A distance profile $h_w=\rho_a(u,w)$ attains the maximum for an
appropriate starting vertex $u$. If $K>0$, then $C(a)=Ka$ on a positive
initial interval. The whole curve is nondecreasing, continuous, and
piecewise linear, with slopes in $\{0,1,\ldots,K\}$.
It is globally $K$-Lipschitz and saturates at
$D(s\!\to\!H)=\tfrac12\max_{u,v}\{S(u,v)-\rho_H(u,v)\}$.
When $K=1$, $C(a)=\min\{a,D(s\!\to\!H)\}$.
\end{theorem}

The directed comparison~\eqref{eq:directed_transfer_main_v10} gives the
cost-curve identity. A finite path calculation gives its complete
representation: with min-plus multiplication
$(A\otimes B)_{uv}=\min_w(A_{uw}+B_{wv})$, set
$B_0=S$ and $B_{m+1}=B_m\otimes U\otimes S$. Then
\begin{equation}\label{eq:joint_cost_envelope}
 \rho_a(u,v)=\min_{0\le m\le K}\{B_m(u,v)+2ma\}.
\end{equation}
Each target segment contributes an additional budget of $2a$; source
segments retain their original budgets. The elementwise minimum $U$
need not be a metric. Each of its entries is taken from one target's
closed distance matrix.

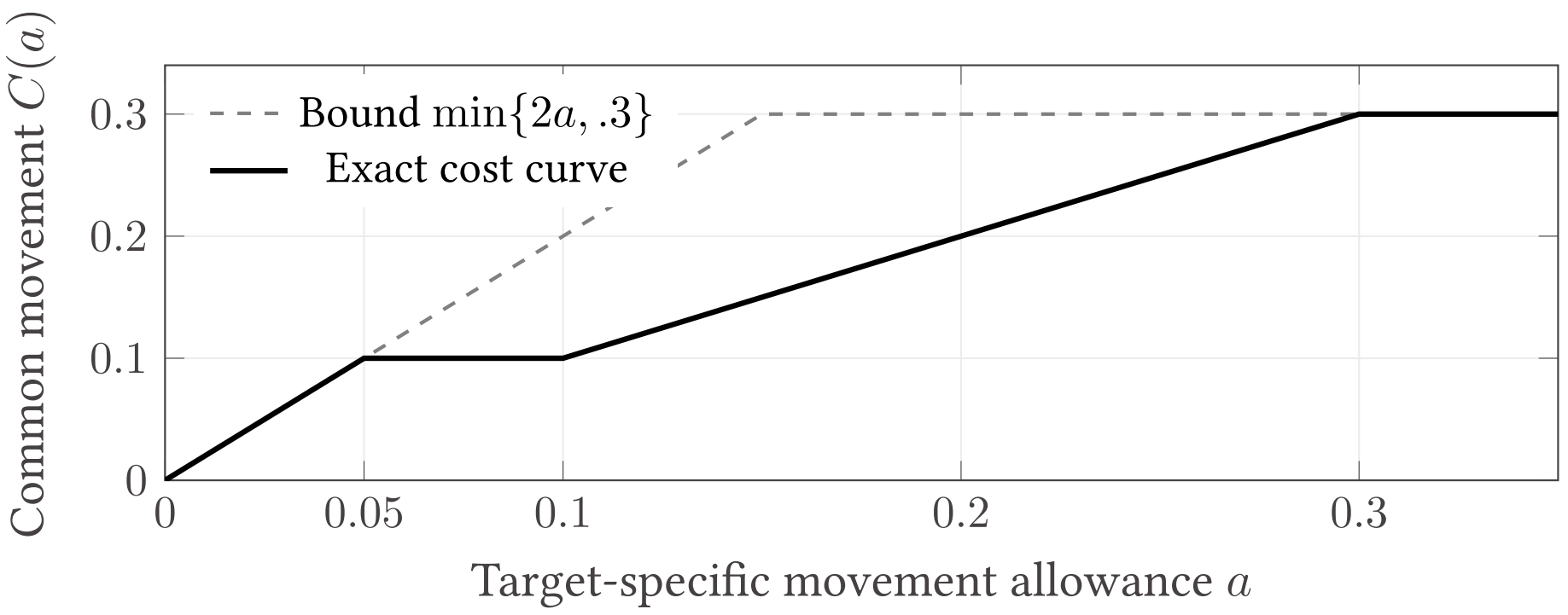
\begin{figure}[t]
\centering
\begin{tikzpicture}
\begin{axis}[
 width=.72\linewidth,height=4.8cm,
 xlabel={Target-specific movement allowance $a$},
 ylabel={Common movement $C(a)$},
 xmin=0,xmax=.35,ymin=0,ymax=.34,
 xtick={0,.05,.1,.2,.3},ytick={0,.1,.2,.3},
 legend style={at={(.02,.98)},anchor=north west,draw=none,font=\small},
 grid=major,grid style={gray!15},
 tick label style={font=\small,/pgf/number format/fixed}]
\addplot[gray,dashed,thick] coordinates {(0,0)(.15,.3)(.35,.3)};
\addlegendentry{Bound $\min\{2a,.3\}$}
\addplot[black,very thick] coordinates {(0,0)(.05,.1)(.1,.1)(.3,.3)(.35,.3)};
\addlegendentry{Exact cost curve}
\end{axis}
\end{tikzpicture}
\caption{Exact cost curve for a six-profile, one-target instance with two
disconnected components: a single edge with source budget $.8$ and target
budget $.2$, and a three-edge source path of budget $.2$ per edge whose target contains
the first and last edges with budget $.1$. Their costs are
$\min\{a,.3\}$ and $\min\{2a,.1\}$, respectively; the overall cost is
their maximum. The resulting curve resumes increasing after a plateau.
Thus $K=2$ determines the initial slope, while the full curve determines
the budget at larger scales. The appendix verifies the formula directly.}
\label{fig:joint_cost_curve}
\end{figure}

After the standard closures and $K$ are available, the entire curve has
an exact representation using $O(Kn^3+n^2)$ arithmetic operations and
$O((K+1)n^2)$ storage. Evaluating the curve at a specified budget requires
one $O(n^3)$ shortest-path computation. A lexicographic shortest-path
algorithm computes $K$ directly
from the labelled union graph. Appendix~\ref{app:joint_adaptation} gives
the algorithms, exact initial and saturation thresholds, and an example
whose slopes increase after a plateau (Figure~\ref{fig:joint_cost_curve}).
The initial amplification factor and saturation cost therefore need not
determine the cost at intermediate budgets.

For example, at $a=.08$ in Figure~\ref{fig:joint_cost_curve}, the exact
uniform budget is $.10$, while $Ka=.16$. A deployment allowance of
$.12$ is therefore certified by the full curve.

Elicitation can reduce the common repair cost by eliminating possible
target constraints. For a fixed predictor, shrinking the candidate family
cannot increase this cost. Its ratio to revealed repair can change in
either direction, since revelation changes the comparison baseline as
well. To certify a deployment budget, we therefore compare the common repair
cost directly with the allowed movement.

\subsection{Source retention with a known target}
\label{sec:source_retention}

For one known target and $h\in\cF_s$, write $r_{s\wedge t}(h)$ for
the minimum movement to $\cF_s\cap\cF_t$. The joint path statistic $K$ now
counts consecutive target runs separated by source edges.

\begin{corollary}[Sharp source-retention bound]
\label{cor:source_runs_main_v16}
\label{thm:source_preservation_main_v11}
For two known graph standards on $n\ge2$ profiles and $h\in\cF_s$,
\begin{equation}\label{eq:source_runs_main_v16}
 r_t(h)\le r_{s\wedge t}(h)\le\min\{1/2,K r_t(h)\},
 \qquad K\le\lfloor n/2\rfloor.
\end{equation}
The coefficient $K$ is exact for these fixed standards, with the local
attainment and finite-budget curve of
Theorems~\ref{thm:joint_amplification}--\ref{thm:joint_cost_curve}.
If every union budget is at least $c>0$, also
$K\le\lfloor\lceil1/c\rceil/2\rfloor$.
The universal factor $\lfloor n/2\rfloor$ is sharp for every $n$,
even with positive tolerances and two fixed positive-definite
Mahalanobis metrics in $\mathbb R^2$.
\end{corollary}
A simple path with $k$ target runs needs at least $k-1$ separating
source edges, giving $2k-1\le n-1$. For sharpness, take ramp scores
$h_i=(i-1)/(n-1)$, a source path with budget $1/(n-1)$ per edge, and
a target matching on its odd edges with a smaller positive budget.
Each matching edge contributes the same deficit; source retention
accumulates all $\lfloor n/2\rfloor$ deficits along the path.
Appendix~\ref{app:source_preservation_v11} gives the construction,
metric realization, and budget-floor argument.

When $K\le1$, retaining the source does not increase the minimum
uniform movement. On a source--target--target path with budgets
$(1/4,1/8,1/8)$, $K=1$, compared with the universal factor $2$.
For $h=(0,1/4,3/4,1)$, the target-only midpoint violates the source
edge, although both optimal repair radii are $1/4$.
Appendix~\ref{app:source_runs_v16} gives the repairs and a bounded
two-dimensional metric realization. Thus repairs with the same optimal radius can differ in which scores
they change, motivating a secondary utility objective (Appendix~\ref{app:repair_selection_v6}).

\section{Adaptation under target uncertainty}
\label{sec:certificates}

We extend the finite-family comparison to a possibly infinite coupled
family, in which one metric determines many edges simultaneously. Let a
nonempty set $\Theta$ index plausible standards on the same pool and
candidate-pair domain $\mathcal P$. Scenario $\theta$ selects edges $E_\theta$ and finite nonnegative budgets $c_e(\theta)$,
with feasible class $\cF_\theta$ and repair cost $r_\theta(h)$ for the given scores.
Threshold fairness sets $E_\theta=\{e:d_e(\theta)\le1\}$ and
$c_e(\theta)=\eps_e(\theta)$; Lipschitz fairness uses its declared pair
domain and budgets $\eps_e(\theta)d_e(\theta)$. Define
\begin{align}
 R_{\rm com}(h;\Theta)&=\min_{z\in\bigcap_{\theta\in\Theta}\cF_\theta}\|z-h\|_\infty,
 \label{eq:common_definition_v6}\\
 R_{\rm rev}(h;\Theta)&=\sup_{\theta\in\Theta}r_\theta(h).
 \label{eq:revealed_definition_v6}
\end{align}
The common cost requires one repair satisfying every plausible standard;
the revealed cost is the worst-case cost of target-specific repair. Both
hold $h$ fixed, unlike the source-class supremum in
\eqref{eq:directed_transfer_main_v10}.

Target uncertainty concerns the identity of a single applicable standard.
Simultaneously binding community standards enter through their intersection.
Continued source compliance is represented by including source edges in
every $E_\theta$ and taking the smaller budget on overlaps. Constant scores ensure feasibility.
A confidence set covering the
true standard with probability $1-\delta$ makes a common repair
feasible under the true standard on that event.

\subsection{Exact common repair for a coupled uncertainty set}
For each pair let $A_e=\{\theta\in\Theta:e\in E_\theta\}$. Construct $G_\cap$ with every edge whose $A_e$ is nonempty and budget
\begin{equation}\label{eq:conditional_budget_main_v6}
 b_e=\inf_{\theta\in A_e}c_e(\theta).
\end{equation}
The infimum ranges over the scenarios in which the edge imposes a constraint.

\begin{theorem}[Common repair characterization]\label{thm:common_main_v6}
The common feasible class equals $\cF_{G_\cap}$, and hence
\[
 R_{\rm com}(h;\Theta)=\tfrac12\max_{u,v}[|h_u-h_v|-\rho_\cap(u,v)]_+.
\]
The midpoint on $G_\cap$ is an optimal common repair. The graph need not be realized by any single $\theta\in\Theta$.
\end{theorem}
\begin{proof}
A common score vector must satisfy every active edge in every scenario. Interchanging these universal quantifiers, it must satisfy each edge at every budget appearing while that edge is active, equivalently at their infimum. This holds even when the infimum is not attained. Apply~\eqref{eq:repair_main}.
\end{proof}

For fixed edge budgets, this is the union of possible edges. Each edge
contributes a requirement to the common repair, including edges from
mutually exclusive scenarios. Their combination can
substantially increase the common cost relative to the revealed cost.

\begin{samepage}
\begin{theorem}[Sharp commitment bound]\label{thm:price_main_v6}
For any nonempty family of nonnegative graph constraints on $n\ge2$ profiles,
\begin{equation}\label{eq:price_main_v6}
 R_{\rm rev}\le R_{\rm com}\le\min\{1/2,(n-1)R_{\rm rev}\}.
\end{equation}
The factor $n-1$ is sharp even for threshold fairness with a known common tolerance, profiles in $\mathbb R^2$, and a full-dimensional convex Frobenius ball of positive-definite $2\times2$ Mahalanobis matrices. The same uncertainty ball works for every $n$, with a fixed strict activation margin at interior matrices that attain the bound. The sharp additive bound is
$R_{\rm com}-R_{\rm rev}\le(n-2)/[2(n-1)]$.
\end{theorem}
\end{samepage}
\begin{proof}
Put $a=R_{\rm rev}$. A scenario-optimal repair implies $|h_u-h_v|\le c_e(\theta)+2a$ on each active edge. Taking the conditional infimum gives $|h_u-h_v|\le b_e+2a$. A shortest path in $G_\cap$ can be chosen simple, with at most $n-1$ edges, so summing gives a score deficit at most $2(n-1)a$. Equation~\eqref{eq:repair_main} proves the factor bound. Constant score $1/2$ proves the range cap. Maximizing $\min\{(n-2)a,1/2-a\}$ gives the additive bound. The construction below proves sharpness.
\end{proof}

\paragraph{Fixed two-dimensional construction.}
Let $x_1=0$ and let $x_{i+1}-x_i$ alternate between $e_1$ and $e_2$.
Use the same matrix uncertainty set for every $n$:
\begin{equation}\label{eq:ball_main_v6}
 \Theta=\{M=M^\top:\|M-2I_2\|_F\le7/5\}.
\end{equation}
Its eigenvalues lie in $[3/5,17/5]$. Odd adjacent pairs have squared
distance $M_{11}$, and even ones $M_{22}$. Activating both orientations
would require displacement at least $\sqrt2>7/5$. Every nonadjacent
displacement has squared norm at least two and hence squared distance
at least $6/5>1$. Thus each scenario induces the odd matching, the even
matching, or no edges; the common graph is the full path.
Interior matrices $\operatorname{diag}(2/3,2)$ and
$\operatorname{diag}(2,2/3)$ realize the two matchings with fixed
squared-distance activation margin $1/3$.

With $h_i=(i-1)/(n-1)$ and $\eps=\gamma/(n-1)$, $0\le\gamma<1$,
disjoint matching edges can each be repaired independently, whereas
the common repair must contract the full endpoint gap. The exact costs are
\begin{equation}\label{eq:matching_radii_main_v9}
 R_{\rm rev}=\frac{1-\gamma}{2(n-1)},\qquad R_{\rm com}=\frac{1-\gamma}{2}.
\end{equation}
Proposition~\ref{prop:alternating_geometry_v9} proves the construction;
setting $\gamma=0$ attains the additive bound. The two interior standards
alone attain these costs. Their fixed activation margins support the query
bounds in Section~\ref{sec:direct_elicitation_v6}; elsewhere in the ball,
targets can approach the activation threshold arbitrarily closely.

\begin{figure}[t]
\centering
\begin{tikzpicture}[v/.style={circle,draw,minimum size=4.6mm,inner sep=1pt,font=\small}, edge/.style={thick}, note/.style={align=center,font=\small}]
\node[note] at (2.2,.9) {Odd matching after revelation};
\node[note] at (9.2,.9) {Even matching after revelation};
\foreach \i in {1,...,5} {
  \node[v] (a\i) at ({1.1*(\i-1)},0) {\i};
  \node[v] (b\i) at ({7+1.1*(\i-1)},0) {\i};
  \node[v] (c\i) at ({3.5+1.1*(\i-1)},-2.05) {\i};
}
\draw[edge] (a1)--(a2);\draw[edge] (a3)--(a4);
\draw[edge] (b2)--(b3);\draw[edge] (b4)--(b5);
\node[note] at (2.2,-.55) {$r_{\rm odd}=.1$};
\node[note] at (9.2,-.55) {$r_{\rm even}=.1$};
\node[note] at (5.7,-1.25) {Common repair before elicitation};
\draw[edge] (c1)--(c2)--(c3)--(c4)--(c5);
\node[note] at (5.7,-2.65) {$R_{\rm com}=.4$};
\end{tikzpicture}
\caption{An $n=5$, $\gamma=.2$ example, with scores $(0,.25,.5,.75,1)$
and edge tolerance $.05$. Each revealed matching costs $.1$ to repair;
a common repair costs $.4$. Repeated judgments on one representative
pair per orientation identify the matching under the response assumptions
in Corollary~\ref{cor:elicitable_sharpness_v9}.}
\label{fig:price_v6}
\end{figure}
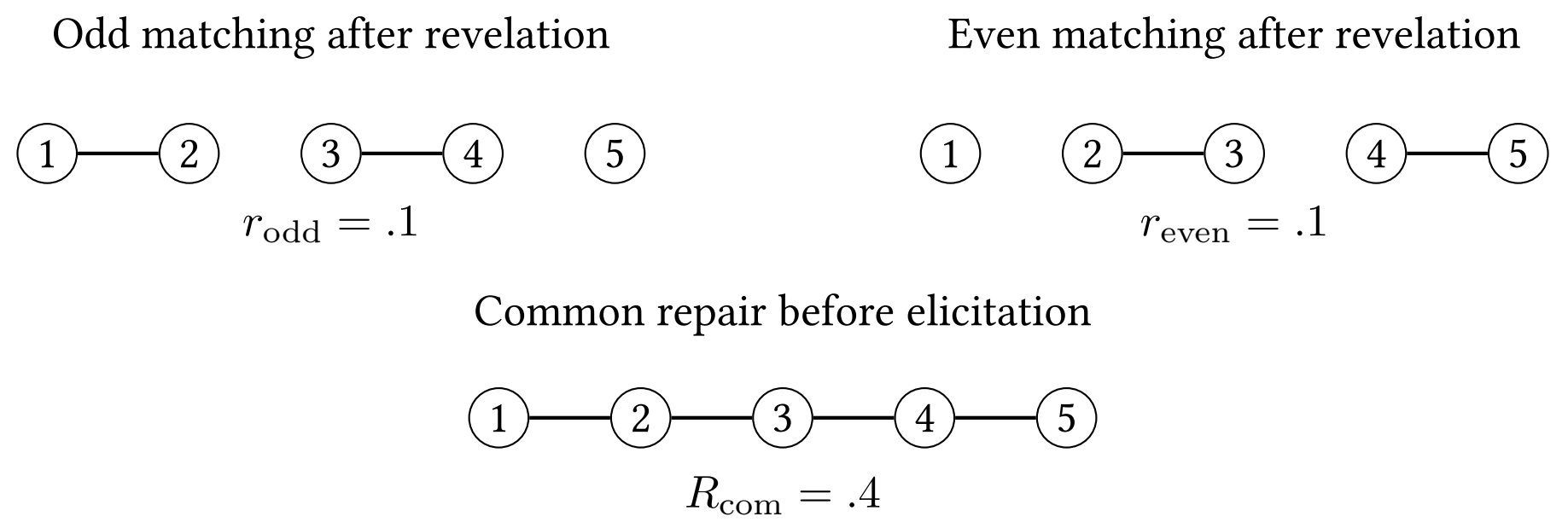

\paragraph{Positive lower bounds on edge budgets.}
If every active budget is at least $c>0$, replace $n-1$ in the factor bound by
$K_c=\min\{n-1,\max\{0,\lceil1/c\rceil-1\}\}$; when $K_c=0$, both costs vanish. A path can create a positive score deficit only if its total budget is below one, so it has fewer than $1/c$ edges. The factor $K_c$ is sharp, and Corollary~\ref{cor:positive_tolerance_v6} gives the sharp additive bound. Thus a fixed positive lower bound on edge budgets bounds the worst-case factor independently of pool size.

\paragraph{Bounded profile domains.}
The alternating pool has diameter $\Theta(n)$. For the full pair domain,
one common tolerance, profiles of diameter $D$
in $\mathbb R^d$, and $M\preceq BI$ throughout the uncertainty set, put
$P=\lfloor(1+2D\sqrt B)^d\rfloor$. Then
\begin{equation}\label{eq:bounded_domain_main_v9}
 R_{\rm com}\le\min\{1/2,\min(n-1,2P-1)R_{\rm rev}\}.
\end{equation}
A path with the fewest edges has no chord, so its alternating vertices must be
more than $1/\sqrt B$ apart. Packing these vertices proves the bound
(Proposition~\ref{prop:bounded_domain_v9}). This caps the factor independently of $n$ even at
zero tolerance. A fixed positive tolerance supplies the further cap above.
The shortcut argument uses both the full pair domain and equal budgets.

\paragraph{Common graph construction.}
For a compact Mahalanobis set and known threshold tolerance, a pair
with difference vector $v$ can be active exactly when
$\min_{M\in\Theta}v^\top Mv\le1$. Compactness ensures attainment at
the threshold. For a Frobenius ball with spectral constraints, this minimization is a
convex conic problem that can be solved separately for each candidate pair. The computational cost
of graph construction can therefore exceed that of the subsequent
shortest-path computation. In the alternating family, the two known pair
directions and the spectral bound determine every possible edge directly.

\section{Elicitation requirements for adaptation}
\label{sec:direct_elicitation_v6}

\paragraph{Movement-budget certification from pairwise responses.}
The elicitation goal is to determine whether target constraints can be
met within a prescribed movement budget and, when needed, to approach
minimum movement.
With fixed known edge tolerances, we query individual pairs or representatives
of known response-equivalence classes, whose pairs have the same response
mean under every candidate standard. Each stream supplies independent
Bernoulli responses with a fixed mean; a pair is constrained exactly when
its mean is at least $1/2$. Simultaneous confidence intervals give a
possible-edge graph with repair cost $r^-$ and a certain-edge graph with
cost $r^+$, satisfying $r^+\le r_{\theta^\star}\le r^-$ on their coverage event.
A movement budget $t$ is feasible with a constructed repair if $r^-\le t$,
impossible if $r^+>t$, and unresolved otherwise. Intervals valid
simultaneously at sample sizes that are powers of two permit adaptive
allocation and stopping. A budget can be resolved without identifying
every near-threshold pair (Theorem~\ref{thm:direct_budget_v6}), adapting
standard thresholding tools~\citep{locatelli2016thresholding} to a
movement-budget decision.

A movement certificate also bounds changes in predictive behavior.
For $h,z\in[0,1]^n$ and $\|z-h\|_\infty\le r\le1$, decisions at threshold
$1/2$ can change only where $|h_i-1/2|\le r$. For any binary labels, the
absolute change in mean Brier loss is at most $2r-r^2$ (Proposition~\ref{prop:repair_utility}).
These are deterministic utility bounds for the repaired pool.

\subsection{Sensitivity to constraint uncertainty}

If $G,G'$ have the same edges and
$\max_e|c_e-c'_e|\le\eta$, then
\begin{equation}\label{eq:transfer_sensitivity_main_v10}
 |r_G(h)-r_{G'}(h)|
 \le\frac12\|\rho_G-\rho_{G'}\|_\infty
 \le\frac12\min\{1,(n-1)\eta\}.
\end{equation}
A simple shortest path has at most $n-1$ edges; its accumulated budget
error yields this sharp bound
(Proposition~\ref{prop:repair_stability}). Once the set of protected pairs is fixed, this bound controls the effect
of tolerance error on the repair radius.
Membership itself can be discontinuous: for two scores $(0,1)$ and
tolerance $0\le\eps<1$, changing their distance from $1-\xi$ to
$1+\xi$, with $0<\xi<1$, changes the repair radius from $(1-\eps)/2$
to zero for arbitrarily small $\xi$. Controlling adaptation error therefore
requires controlling edge activation near the threshold. For a
Lipschitz rule on a fixed pair domain, metric changes instead perturb
edge budgets and the same sensitivity bound applies to those perturbations.

Jointly valid distance and tolerance bands give safe and relaxed
graphs bracketing the true cost (Theorem~\ref{thm:uncertain_repair}),
which can certify or rule out a movement budget.

\subsection{Tolerance precision for known target graphs}
\label{sec:tolerance_geometry}

Once the target graph is known, the tolerance information needed for
adaptation depends on the paths that determine its repair cost. This
dependence holds for arbitrary graph topology and also when source
constraints remain binding. Let $S$ be a known source graph with fixed
nonnegative budgets, let $T$ be a known target graph, and suppose
$h\in\cF_S$. Every target edge has a common unknown tolerance
$\eps\in[0,1]$. Write
$r(\eps)=\operatorname{dist}_\infty(h,\cF_S\cap\cF_{T,\eps})$.
Retain parallel source and target edges on overlaps; this is equivalent to
using the smaller of their budgets.

For a simple path $P$ in this multigraph, let $A_P$ be the sum of its
source-edge budgets and $k_P$ its number of target edges. Let
$H_P=|h_u-h_v|$ for its endpoints. A \emph{critical path} at $a$ attains
the positive maximum in the following expression.

\begin{theorem}[Critical paths and tolerance sensitivity]
\label{thm:tolerance_path_geometry}
The exact repair radius under source retention is
\begin{equation}\label{eq:tolerance_path_envelope}
 r(\eps)=\frac12\max\bigl(\{0\}\cup
 \{H_P-A_P-k_P\eps:P\text{ is a simple path}\}\bigr).
\end{equation}
Consequently, $r$ is nonincreasing, convex, and piecewise affine.
If $r(a)>0$, define
\[
 \kappa_{\rm tol}(a)=\min\{k_P:H_P-A_P-k_Pa=2r(a)\};
\]
set $\kappa_{\rm tol}(a)=0$ when $r(a)=0$. Then, for
$a\le x\le y$,
\begin{equation}\label{eq:tolerance_critical_sensitivity}
 0\le r(x)-r(y)\le\frac{\kappa_{\rm tol}(a)}2(y-x).
\end{equation}
This coefficient is exact locally to the right of $a$:
$r'_+(a)=-\kappa_{\rm tol}(a)/2$. If $r(a)>0$, source feasibility
implies $1\le\kappa_{\rm tol}(a)\le n-1$.
The radius, one critical path with the fewest target edges, and this
coefficient can be computed in
$O((|E_S|+|E_T|+n)\log n)$ time by one multi-source Dijkstra search
with target-edge counts used to break distance ties.
\end{theorem}

The path formula follows from the exact graph repair identity. A maximum
of finitely many affine functions has right derivative equal to the
largest slope among its active functions, giving the \emph{fewest}
target edges at a tie. Convexity then gives the bound throughout the
remaining tolerance range. Appendix~\ref{app:tolerance_geometry} proves
the statement and specifies the search. Thus the relevant tolerance
sensitivity can be much smaller than the worst-case $(n-1)/2$; fixed
source edges contribute to path length but require no tolerance inference.

\paragraph{A movement certificate from one critical path.}
The same reasoning permits different tolerances on different target
edges. Suppose jointly valid bands $[\ell_e,u_e]$ are available, and
let $r_\ell$ be the radius of the safe graph using all lower endpoints.
If $r_\ell>0$, any critical path $P$ of that graph gives
\begin{equation}\label{eq:tolerance_path_width}
 0\le r_\ell-r_\star
 \le\frac12\sum_{e\in P\cap E_T}(u_e-\ell_e).
\end{equation}
The safe repair is feasible under the true specification. Its excess
movement is therefore at most $\eta$ if the displayed path-width sum is
at most $2\eta$, or its movement is at most $(1+\alpha)r_\star$ if that
sum is at most $2\alpha r_\ell/(1+\alpha)$. If $r_\ell=0$, return $h$.
Shared community tolerances contribute once for each occurrence on the
path. With bands valid across inspected sample sizes, one may query
tolerance streams on the current critical path, recompute after
tightening the bands, and stop at either certificate. The critical path
can change, so this rule can visit several paths; the guarantee concerns
its returned repair rather than optimal query allocation.

\paragraph{Response complexity and a local lower bound.}
Fix an acceptance level $q\in(0,1)$ and known slope $\lambda>0$.
A query displaying gap $g\in[0,1]$ receives a fresh Bernoulli response
whose conditional mean, given the past and selected gap, is
\begin{equation}\label{eq:tolerance_oracle_geometry}
 p_\eps(g)=\sigma\bigl(\operatorname{logit}(q)+\lambda(\eps-g)\bigr),
 \qquad \sigma(s)=(1+e^{-s})^{-1}.
\end{equation}
Here $\eps$ is the $q$-level acceptance cutoff for a fixed respondent
population and presentation protocol. The target graph is known;
this response assumption supplies tolerance information.

\begin{theorem}[Tolerance response complexity on known target graphs]
\label{thm:tolerance_general_queries}
Suppose $\eps\in[a,b]\subseteq[0,1]$, $a<b$, and $r_{\min}=r(b)>0$.
Let $0<\alpha\le1/4$, $0<\delta<1/2$, and
\[
 v=\min_{|s|\le\lambda(b-a)/2}
 \sigma(\operatorname{logit}(q)+s)
 [1-\sigma(\operatorname{logit}(q)+s)]>0.
\]
Under~\eqref{eq:tolerance_oracle_geometry}, at most
\begin{equation}\label{eq:tolerance_general_upper}
 N=\left\lceil
 \frac{\kappa_{\rm tol}(a)^2\log(2/\delta)}
 {2\lambda^2v^2\alpha^2r_{\min}^2}
 \right\rceil
\end{equation}
responses at the single gap $g=(a+b)/2$ suffice to return, with
probability at least $1-\delta$, a source-retaining repair feasible under
the true target and having movement at most $(1+\alpha)r(\eps)$.

Conversely, suppose $r_0=r(\eps_0)>0$ and
$r(\eps)=r_0-k(\eps-\eps_0)/2$ throughout
$[\eps_0,\eps_1]\subseteq[0,1]$, where $k>0$ and
$\eps_1=\eps_0+2\alpha r_0/k$.
Any almost surely terminating procedure achieving the same guarantee at
both endpoints must satisfy, for $i\in\{0,1\}$,
\begin{equation}\label{eq:tolerance_general_lower}
 \mathbb E_iN_{\rm tol}\ge
 \frac{2k^2\operatorname{kl}(1-\delta,\delta)}
 {\lambda^2\alpha^2r_0^2}.
\end{equation}
This lower bound permits adaptive gap choices and other observations
whose conditional laws coincide under the two tolerances.
Here $\operatorname{kl}$ denotes Bernoulli relative entropy.
\end{theorem}

For the upper bound, invert a Hoeffding interval and repair at its lower
tolerance endpoint; Theorem~\ref{thm:tolerance_path_geometry} controls
excess movement. For the lower bound, feasibility at $\eps_0$ forces
movement at least $r_0$, whereas success at $\eps_1$ forces movement at
most $(1-\alpha^2)r_0$. The two tolerances must therefore be distinguished.
Appendix~\ref{app:tolerance_geometry} gives the complete proofs.
For $\delta\le1/10$, on an affine prior interval containing the two
lower-bound alternatives and having comparable positive repair radii,
the bounds match in target-edge count, movement scale, $\alpha$, and
$\delta$ at fixed $q,\lambda$. The relevant slope can change when
another path becomes critical.
If $r(b)=0$, the additive certificate~\eqref{eq:tolerance_path_width}
still applies, but the stated uniform relative bound does not.

\subsection{Graph identification and tolerance precision}
The alternating family of Proposition~\ref{prop:alternating_geometry_v9}
distinguishes information about active constraints from precision in
their tolerances. Let $n\ge3$, $\Delta=1/(n-1)$, and $h_i=(i-1)\Delta$.
The profile representation and matrix ball are known; the target matrix
is unknown, and its graph is an alternating matching or empty.

\paragraph{Graph identification from similarity queries.}
Odd adjacent pairs have squared distance $M_{11}$ and even ones $M_{22}$;
the spectral bound excludes every nonadjacent pair. Under the common
logistic similarity model $p_e=\sigma(\beta[1-d_e])$ with known
slope $\beta>0$, repeated independent judgments on $(1,2)$
and $(2,3)$ therefore determine whether each orientation is active. At the two interior
matrices, let $\kappa_0=1-\sqrt{2/3}$ and
$a_0=\tfrac12\tanh(\beta\kappa_0/2)$. Hoeffding intervals identify the
graph with probability at least $1-\delta$ using at most
$2(\lfloor2a_0^{-2}\log(4/\delta)\rfloor+1)$ responses, independently
of $n$. For every matrix in the ball, the procedure is correct or
abstains on its confidence event; its sample complexity depends on
the activation margin.
With known tolerance it returns the exact optimal repair
(Corollary~\ref{cor:elicitable_sharpness_v9}). Known geometry permits
graph construction and repair in $O(n)$ time, without all-pairs searches.

\paragraph{Excess movement from tolerance uncertainty.}
For a correctly identified nonempty matching, the true repair radius is
$r_\eps=[\Delta-\eps]_+/2$. A valid tolerance interval
$0\le\ell\le\eps\le u$ gives a safe repair at $\ell$ with
\begin{equation}\label{eq:tolerance_excess_main_v11}
 0\le r_\ell-r_\eps\le(u-\ell)/2.
\end{equation}
Before identifying the matching, the common path has radius
$R_\eps=\tfrac12[1-(n-1)\eps]_+$, whose tolerance sensitivity below
$\Delta$ is $(n-1)/2$. Thus the general path-length sensitivity above becomes $1/2$ after identifying the graph
(Proposition~\ref{prop:tolerance_interval_v11}).

For tolerance inference, use the fixed-slope acceptance model
\eqref{eq:tolerance_oracle_geometry}. The similarity response model
depends only on the metric, so its responses carry no tolerance
information. The general graph result above supplies the dependence on
repair sensitivity; the matching family makes the resulting rate explicit.

\begin{theorem}[Tolerance queries for constant-factor and near-optimal repair]
\label{thm:tolerance_tradeoff_main_v11}
For $n\ge3$, suppose the unknown common tolerance lies in $[0,\Delta/2]$.
At either of the two interior target matrices, identifying the matching and
equalizing its active pairs uses no tolerance responses, is feasible under the true standard,
and returns movement $\Delta/2\le2r_\eps$ on the graph-confidence event.
Common repair with the true tolerance requires at least
$(n-1)/2$ times as much movement.

For $0<\alpha\le1/4$ and $0<\delta\le1/10$, under~\eqref{eq:tolerance_oracle_geometry},
at those same matrices, a repair feasible under the true standard with movement
at most $(1+\alpha)r_\eps$ can
instead be returned with probability at least $1-\delta$ using
\begin{equation}\label{eq:joint_tolerance_rate_main_v11}
 O\!\left(\alpha^{-2}(n-1)^2\log(1/\delta)\right)
\end{equation}
additional tolerance responses, for fixed $q,\lambda$. Conversely, even
with the matching and metric supplied, any almost surely terminating
procedure achieving this guarantee for every $\eps\in[0,\Delta/2]$
requires, at each of two tolerances,
\begin{equation}\label{eq:tolerance_lower_main_v11}
 \mathbb E N_{\rm tol}\ge
 \frac{8(n-1)^2\operatorname{kl}(1-\delta,\delta)}{\lambda^2\alpha^2}.
\end{equation}
Here $\operatorname{kl}$ is Bernoulli relative entropy. The bounds match
in $n$, $\alpha$, and $\delta$ at fixed $q,\lambda$.
\end{theorem}
The general tolerance upper bound applies with
$\kappa_{\rm tol}=1$ and $r_{\min}=\Delta/4$.
The local lower bound likewise has order
$\alpha^{-2}\Delta^{-2}\log(1/\delta)$ on the matching's affine repair
curve. Appendix~\ref{app:tolerance_tradeoff_v11} gives the explicit
lower-bound alternatives establishing the displayed constant, together
with the graph-confidence allocation and the $2$-approximation proof.

The quadratic response cost reflects the precision needed to estimate a
tolerance of order $1/n$ under a fixed noise scale in score-gap units.
Graph identification already permits a constant-factor repair; approaching
the optimal movement requires this finer tolerance estimate. Allowing the
response slope to increase with $n$ changes the noise scale and hence the cost.

\section{Eliciting standards and evaluating predictive utility}
\label{sec:experiments}

We study how respondents specify similarity and acceptable score gaps on
COMPAS~\citep{compas}, then compare numerical compliance and predictive
utility under different training tolerances and score compression.
Appendix~\ref{app:experiments} gives the study tasks, estimators, training
procedure, and evaluation protocol.

\paragraph{Similarity and tolerance judgments.}
The similarity study uses two defendant populations, African-American and
Caucasian, with ten annotators and 500 pairs per population. Profile cards
show demographic and criminal-history information with race omitted.
Fleiss' $\kappa$ is $.549$ and $.512$. We fit a Mahalanobis matrix for
each population to majority-vote similarity labels.

The tolerance study recruits thirty Black/African-American respondents
and thirty White/Caucasian respondents through Prolific. Each respondent
judges 200 pairs from the corresponding fitted unit ball, stratified
into five score-gap bins, for 12,000 responses in total. A
logistic-regression scorer supplies the displayed risks. Respondents
judge whether each score gap is acceptable for similarly situated
defendants. Table~\ref{tab:phase2} reports two summaries: a fitted
logistic 50\% acceptance point and a cumulative-grid cutoff requiring
90\% acceptance among observations below the cutoff.

The bootstrap interval for the difference in grid medians
(White/Caucasian minus Black/African-American) is $[-.035,.075]$,
with permutation $p=.10$ and Cliff's $\delta=.16$. Mann--Whitney $p$
is $.05$ for logistic summaries and $.14$ for grid summaries.
The grid interval spans zero, so the two panel medians alone do not
establish a population difference. The fitted logistic medians
$(.039,.063)$ provide the numerical tolerance values used below.

\begin{table}[!htbp]
\centering\small
\caption{COMPAS tolerance summaries, thirty respondents per community and 200 pairs each. Grid and logistic columns summarize different estimands. Acceptance is the overall percentage of judgments marking the displayed score gap acceptable.}
\label{tab:phase2}
\begin{tabular}{lccccc}
\toprule
Respondent panel & Grid median & Logistic median & Grid IQR & $\kappa$ & Acceptance\\
\midrule
Black/African-American & .035 & .039 & $[.020,.100]$ & .40 & 42\%\\
White/Caucasian & .060 & .063 & $[.035,.178]$ & .43 & 50\%\\
\bottomrule
\end{tabular}
\end{table}

\paragraph{Training tolerances and score compression.}
Table~\ref{tab:common_protocol_v5} compares two soft-penalty training
procedures at three tolerance settings, together with compression of
the Lipschitz-U predictor. All rows use the same evaluation tolerances
$(.039,.063)$ and sampled pairs within each of five data splits.
The benchmark representation replaces charge degree and sex in the
metric-fitting coordinates with two COMPAS decile scores. The matrices
therefore specify numerical quadratic forms for this comparison;
their elicited interpretation applies to the fitting representation.

For Lipschitz training, reducing the tolerances from $(.10,.10)$ to
$(.039,.063)$ lowers strict violation from $.2212$ to $.0339$, while
accuracy falls from $.6809$ to $.6717$ and Brier loss increases from
$.2113$ to $.2183$. Threshold training shows the same direction of
change. The two procedures differ in pair sampling as well as penalty
form, so their rows compare the complete training procedures.

Compression maps each score to
$h_\alpha(x)=\tfrac12+\alpha(h(x)-\tfrac12)$, with $0<\alpha\le1$.
It preserves decisions at threshold $0.5$ while shrinking every score
gap by $\alpha$. At $\alpha=.1$, observed strict violation is zero and
accuracy remains $.6809$, while Brier loss rises to $.2439$ and log
loss to $.6809$. Brier and log loss are proper scoring rules~\citep{gneiting2007strictly}
and distinguish these predictors despite their identical threshold
decisions. Together, the comparisons motivate reporting predictive
loss alongside compliance and the movement objective studied here.

\begin{table}[!htbp]
\centering\scriptsize
\setlength{\tabcolsep}{3pt}
\caption{Training and score compression under common evaluation: means $\pm$ standard deviations across five data splits and training runs. U, H, and M denote training tolerances $(.10,.10)$, $(.039,.063)$, and $(.039,.039)$. Compression uses the Lipschitz-U base. All rows are evaluated at $(.039,.063)$; lower violation, Brier, and log loss are better. The $\alpha=1$ row repeats the base for reference.}
\label{tab:common_protocol_full_v5}
\label{tab:common_protocol_v5}
\begin{tabular}{lrrrr}
\toprule
Procedure & Accuracy & Strict viol. & Brier & Log loss\\
\midrule
Lipschitz-U & $.6809\pm.0060$ & $.2212\pm.0214$ & $.2113\pm.0019$ & $.6122\pm.0048$\\
Lipschitz-H & $.6717\pm.0055$ & $.0339\pm.0044$ & $.2183\pm.0018$ & $.6275\pm.0044$\\
Lipschitz-M & $.6664\pm.0065$ & $.0187\pm.0027$ & $.2220\pm.0016$ & $.6354\pm.0040$\\
Threshold-U & $.6793\pm.0084$ & $.3116\pm.0229$ & $.2096\pm.0028$ & $.6087\pm.0068$\\
Threshold-H & $.6783\pm.0067$ & $.1614\pm.0050$ & $.2119\pm.0020$ & $.6141\pm.0056$\\
Threshold-M & $.6755\pm.0059$ & $.1036\pm.0094$ & $.2140\pm.0023$ & $.6187\pm.0061$\\
\midrule
Compression, $\alpha=1.00$ & $.6809\pm.0060$ & $.2212\pm.0214$ & $.2113\pm.0019$ & $.6122\pm.0048$\\
Compression, $\alpha=0.75$ & $.6809\pm.0060$ & $.0954\pm.0220$ & $.2163\pm.0016$ & $.6234\pm.0036$\\
Compression, $\alpha=0.50$ & $.6809\pm.0060$ & $.0204\pm.0060$ & $.2243\pm.0012$ & $.6410\pm.0024$\\
Compression, $\alpha=0.25$ & $.6809\pm.0060$ & $.0038\pm.0005$ & $.2356\pm.0006$ & $.6642\pm.0013$\\
Compression, $\alpha=0.10$ & $.6809\pm.0060$ & $.0000\pm.0000$ & $.2439\pm.0003$ & $.6809\pm.0005$\\
\bottomrule
\end{tabular}
\end{table}

\section{Related work}
\label{sec:related}

\paragraph{Fairness standards and elicitation.}
Individual fairness specifies similar treatment through a similarity rule~\citep{dwork2012fairness}. Human-supervised metric learning~\citep{wang2019empirical}, metric approximation~\citep{ilvento2019metric}, and learned invariances~\citep{mukherjee2020two} estimate its geometry. Other frameworks elicit pairwise constraints directly~\citep{jung2021algorithmic,bechavod2020metricfree}. Our setting separates protected pairs from their acceptable score gaps and studies the movement required when either changes. Individual-fairness training methods~\citep{yurochkin2021sensei} optimize predictors under specified constraints; finite-pool repair gives the minimum output movement for a supplied predictor and graph.

\paragraph{Uniform approximation and graph repair.}
Lipschitz envelopes characterize uniform approximation~\citep{mcshane1934extension,mustata2007}. Applying them to a graph's shortest-path closure gives the repair and directed-distance identities used here. Fairness post-processing also includes squared-change projection and graph regularization~\citep{petersen2021postprocessing}. We use the envelope construction to compute optimal uniform repairs. The main characterization concerns the additional movement caused by retaining source constraints and combining target possibilities: it gives an exact coefficient for each fixed family, predictors attaining the bound, and the complete finite-budget cost curve.

\paragraph{Error bounds and commitment under uncertainty.}
Hoffman's theorem bounds distance to linear-inequality feasibility by a residual~\citep{hoffman1952}. Sharp constants admit general variational characterizations~\citep{aze2002hoffman}, and relative Hoffman bounds incorporate a reference polyhedron that is already satisfied~\citep{pena2021hoffman}. Our result evaluates a fixed-right-hand-side relative error constant for graph standards. Specifically, take the shortest-path closure of each target and stack its ordered-pair inequalities into $Ah\le b$. On the source class $P=\cF_s$, the residual is exactly $\|(Ah-b)_+\|_\infty=2R(h)$, so the optimal fixed-$b$ coefficient is $K/2$. The relative constant of \citet{pena2021hoffman} is uniform over feasible right-hand sides and can be strictly larger. Appendix~\ref{app:joint_hoffman} gives the formal comparison. The target-run formula evaluates this structured constant through shortest paths, constructs scores attaining the bound, and extends the comparison to every finite residual budget.

An edge-residual bound depends only on the union inequalities. The comparison of the common repair cost with the largest target-specific repair cost also depends on which edges belong to the same target. For an unconstrained source and a path with total budget below one, assigning all edges to one target gives $K=1$, while splitting the same edges into two alternating target matchings gives $K=n-1$. The target-run formula makes this grouping explicit, and the cost curve resolves its effect beyond the local error constant.

Commitment before uncertainty resolves and adjustment afterward are established robust-optimization notions~\citep{bental2004adjustable,bertsimas2010power}. For graph constraints, the common feasible class is obtained by taking each possible edge at its smallest active budget. A path-residual argument gives the factor $n-1$ between common and revealed repair; our two-dimensional metric family attains it within a fixed convex uncertainty ball. The joint characterization connects this comparison to source retention, where one target permits the smaller sharp factor $\lfloor n/2\rfloor$.

\paragraph{Queries and certificates.}
Thresholding bandits identify response means relative to a threshold~\citep{locatelli2016thresholding}. Our query analysis uses concentration and sequential change-of-measure tools~\citep{kaufmann2016complexity}, with the repair objective determining the precision required of the responses. Confidence graphs certify movement-budget decisions; critical paths translate tolerance uncertainty into excess repair movement. The matching construction shows how the tolerance-query cost can exceed the cost of identifying the protected pairs.

Distributional metric-fair learning and certification~\citep{rothblum2018probably,wicker2023certification} concern guarantees beyond a supplied pool. Proper scoring rules evaluate the predictive quality of repaired probabilities~\citep{gneiting2007strictly}, complementing the movement objective studied here.

\section{Discussion}
\label{sec:conclusion}

Adapting scores to a newly elicited standard can require additional movement to retain source constraints or satisfy all candidate targets. The joint cost curve characterizes both requirements for a finite family of graph standards. Its smallest amplification coefficient counts target-standard runs along shortest union paths. At larger movement budgets, different paths can determine the cost. Because the coefficient depends on which labelled paths are exactly shortest, small budget perturbations can change its value. The full curve specifies how far the initial linear regime extends.

The geometric realizations identify when these costs can grow with the pool. The sharp families use growing diameter and shrinking or zero budgets. A positive budget floor caps both factors, and bounded geometry caps commitment cost for a complete pair domain with a common tolerance. Tighter geometric bounds and efficient representations of large target families would extend the range of instances for which these comparisons can be computed directly. Infinite coupled metric sets also require attention to graph construction, since each candidate pair may entail an optimization problem.

The elicitation analysis distinguishes uncertainty about protected pairs from uncertainty about their tolerances. Movement-budget certificates can resolve the adaptation decision while some pairs remain uncertain. For a known topology, critical paths quantify the tolerance precision required by a repair objective. The logistic response bounds assume a fixed respondent population, presentation design, and response slope in score-gap units; the relative-accuracy guarantee uses a positive lower bound on the optimal repair radius. The explicit matching example additionally uses known response classes and interior activation margins. Learning those classes and allocating queries jointly across topology and tolerance are natural next steps.

Exact repair optimizes scores on the declared pool. Applying its guarantees to new profiles requires an extension rule and assumptions on the deployment distribution. Within the pool, multiple repairs can attain the same maximum movement while differing in predictive loss. Selecting among them using a secondary utility objective, and comparing their movement with that of a parameterized predictor, would connect the geometric optimum more directly to model adaptation.

Community definitions, representations, panel composition, and score semantics determine the meaning of the constraints. Within-community protection covers the declared comparisons; additional cross-community obligations require explicit rules. Elicitation can resolve uncertainty about these rules, while reconciliation of simultaneously binding obligations remains a normative choice. Reporting the standard alongside its repair cost allows readers to assess how these choices affect the required score changes.

\section*{Acknowledgments and disclosures}
I am employed by Microsoft. This work received no funding or other support. I have no other competing interests to declare.

Language models assisted literature comparison, mathematical development and checking, and exposition.

\clearpage
\appendix
\section{Exact graph repair and optimal selections}
\label{app:repair}

We obtain finite-pool repairs by applying classical Lipschitz distance
envelopes~\citep{mcshane1934extension,mustata2007} to the shortest-path
closure of the constraints. The resulting radius and envelope formulas
support the source-retention and uncertainty comparisons below.

\subsection{Exact minimum-change repair on a finite pool}
\label{app:exact_pool_repair_v3}

Fix a finite vertex set $V=\{1,\ldots,n\}$, $n\ge1$, an undirected graph
$G=(V,E)$, and nonnegative finite budgets $c_{uv}$ on its edges. Write
\[
 \mathcal F_G=\{z\in[0,1]^n:
 |z_u-z_v|\le c_{uv}\ \text{for every }(u,v)\in E\}.
\]
Threshold constraints use $c_{uv}=\epsilon$ on the selected neighborhood
edges; local Lipschitz constraints use
$c_{uv}=\epsilon d(x_u,x_v)$ on those edges. The construction therefore
respects the specified constraint graph, including decisions to omit
edges or to include cross-community edges.

Let $d_G(u,v)$ be shortest-path distance with edge lengths $c_{uv}$,
with $d_G(u,v)=+\infty$ for disconnected vertices and $d_G(u,u)=0$.
Define the range-capped path pseudometric
\[
 \rho_G(u,v)=\min\{1,d_G(u,v)\}.
\]
Zero-length edges are allowed. The shortest-path triangle inequality and
$\min\{1,a+b\}\le\min\{1,a\}+\min\{1,b\}$ show that $\rho_G$ is a
pseudometric. Path summation and the score range give the equivalence
\begin{equation}\label{eq:graph_path_equivalence_v3}
 z\in\mathcal F_G
 \quad\Longleftrightarrow\quad
 |z_u-z_v|\le\rho_G(u,v)\quad\text{for all }u,v\in V.
\end{equation}
In the reverse direction, use
$\rho_G(u,v)\le d_G(u,v)\le c_{uv}$ on each edge.

\begin{proposition}[Exact finite-pool uniform repair radius]
\label{prop:exact_pool_repair_v3}
For $h\in[0,1]^n$, let
\[
 \Delta_G(h)=\max_{u,v\in V}
 [|h_u-h_v|-\rho_G(u,v)]_+.
\]
Then
\begin{equation}\label{eq:exact_pool_repair_v3}
 r_G(h):=\min_{z\in\mathcal F_G}\|z-h\|_\infty
 =\frac12\Delta_G(h).
\end{equation}
One minimizer is obtained from the lower distance envelope
\[
 L_u=\min_{v\in V}\{h_v+\rho_G(u,v)\},
 \qquad
 z_u^\star=\min\{1,L_u+\Delta_G(h)/2\}.
\]
\end{proposition}

\begin{proof}
For any $z\in\mathcal F_G$, the triangle inequality and
\eqref{eq:graph_path_equivalence_v3} imply
\[
 |h_u-h_v|\le 2\|z-h\|_\infty+\rho_G(u,v),
\]
so every feasible repair has error at least $\Delta_G(h)/2$.

The choice $v=u$ and nonnegativity give $0\le L_u\le h_u\le1$.
Moreover,
\[
 0\le h_u-L_u
 =\max_v\{h_u-h_v-\rho_G(u,v)\}\le\Delta_G(h).
\]
The pseudometric triangle inequality gives
$|L_u-L_v|\le\rho_G(u,v)$. Adding the same constant preserves these
differences, and clipping to $[0,1]$ cannot enlarge a difference.
Consequently $z^\star\in\mathcal F_G$. Before clipping,
$L_u+\Delta_G(h)/2-h_u$ lies in
$[-\Delta_G(h)/2,\Delta_G(h)/2]$; clipping cannot increase its distance
from $h_u\in[0,1]$. This attains the lower bound.
\end{proof}

\paragraph{Shortest-path computation of an optimal repair.}
\label{app:envelope_algorithm_v5}
For a fixed score vector $h$, the envelope and repair radius can be
computed by one shortest-path search. In fact,
\[
 L_u=\min_v\{h_v+d_G(v,u)\},\qquad
 r_G(h)=\frac12\max_u(h_u-L_u).
\]
To prove the first equality, observe that a capped-distance candidate
with $d_G(v,u)\ge1$ has value $h_v+1\ge1$, whereas the candidate $v=u$
has value $h_u\le1$. Replacing these candidates by their uncapped values,
including $+\infty$ across components, therefore leaves the minimum
unchanged. For the second equality, symmetry of $\rho_G$ and the zero
deficit at $u=v$ give
\[
 \max_u(h_u-L_u)
 =\max_{u,v}\{h_u-h_v-\rho_G(u,v)\}
 =\Delta_G(h).
\]
Add a source vertex with an outgoing arc of length $h_v$ to every $v$.
Its shortest-path distance to $u$ is exactly $L_u$. All edge and arc
lengths are nonnegative, so Dijkstra's algorithm with a binary heap
computes all $L_u$ in $O((|E|+n)\log n)$ time and $O(|E|+n)$ space
for $n\ge2$; the singleton case is immediate. Equivalently, initialize a
multi-source search with tentative distance $h_v$ at every vertex.
Linear scans then recover the radius and the shifted-envelope minimizer.
The directed class comparison below uses the two all-pairs distance matrices.

\begin{corollary}[Directional transfer for local graph constraints]
\label{cor:graph_transfer_v3}
Let source and target graphs have the same vertex set, with feasible
classes $\mathcal F_s,\mathcal F_t$ and range-capped path pseudometrics
$\rho_s,\rho_t$. Then
\begin{equation}\label{eq:directed_graph_repair_v3}
 \sup_{h\in\mathcal F_s}\inf_{z\in\mathcal F_t}
 \|h-z\|_\infty
 =\frac12\max_{u,v\in V}[\rho_s(u,v)-\rho_t(u,v)]_+.
\end{equation}
This is the worst-case minimum uniform repair cost over source-feasible scores.
\end{corollary}

\begin{proof}
For $h\in\mathcal F_s$, equation~\eqref{eq:graph_path_equivalence_v3}
and Proposition~\ref{prop:exact_pool_repair_v3} give the upper bound.
If the maximum on the right is zero, nonnegativity gives equality.
Otherwise, choose a maximizing pair $(u,v)$ and put
$h_w=\rho_s(w,u)$. The triangle inequality gives
$h\in\mathcal F_s$, with
$|h_u-h_v|=\rho_s(u,v)$. The exact repair identity now gives the matching
lower bound.
\end{proof}

\subsection{Optimal repair selection and component preservation}
\label{app:repair_selection_v6}

The shifted lower envelope proves attainment but can move an isolated
score. A symmetric choice uses both distance envelopes:
\begin{equation}\label{eq:midpoint_repair_v6}
 L_u=\min_v\{h_v+\rho_G(u,v)\},\qquad
 U_u=\max_v\{h_v-\rho_G(u,v)\},\qquad
 M_u(h)=\frac{L_u+U_u}{2}.
\end{equation}

\begin{proposition}[Midpoint repair and the set of optimal repairs]
\label{prop:repair_selection_v6}
The midpoint $M(h)$ is feasible and attains the minimum uniform radius
$r_G(h)$. On each connected component $C$ it attains that component's
minimum repair radius. In particular, it leaves isolated vertices and
already feasible components unchanged. For any $r\ge r_G(h)$, let
\[
 A_u(r)=\max\{0,U_u-r\},\qquad
 B_u(r)=\min\{1,L_u+r\}.
\]
The entire set of repairs with movement at most $r$ is
\begin{equation}\label{eq:all_repairs_v6}
 \mathcal C_r
 :=\{z\in\mathcal F_G:\|z-h\|_\infty\le r\}
 =\{z\in\mathcal F_G:A(r)\le z\le B(r)\}.
\end{equation}
Both $A(r)$ and $B(r)$ belong to $\mathcal C_r$ and are its
coordinatewise minimum and maximum. Intermediate coordinates cannot
in general be selected independently of the graph constraints.
\end{proposition}

\begin{proof}
Both envelopes are $1$-Lipschitz under $\rho_G$ and satisfy
$0\le L\le h\le U\le1$, so their midpoint is feasible. Writing
$\Delta=2r_G(h)$, the deficits $a_u=h_u-L_u$ and $b_u=U_u-h_u$
both lie in $[0,\Delta]$. Thus
$|M_u(h)-h_u|=|b_u-a_u|/2\le\Delta/2$; the lower bound in
Proposition~\ref{prop:exact_pool_repair_v3} proves optimality.
Candidates across disconnected components
have capped distance one and cannot improve either envelope.
Applying the same argument within a component proves componentwise
optimality and shows that feasible components remain unchanged.

If $z\in\mathcal C_r$, its Lipschitz constraints give
$z_u\ge\max_v\{h_v-r-\rho_G(u,v)\}=U_u-r$ and
$z_u\le\min_v\{h_v+r+\rho_G(u,v)\}=L_u+r$.
Conversely, these bounds imply $h-r\le z\le h+r$ because
$U\ge h$ and $L\le h$. Clipping envelopes against constants
preserves feasibility. Finally, the triangle inequality gives
\[
 U_u-L_u
 =\max_{v,w}\{h_v-h_w-\rho_G(u,v)-\rho_G(u,w)\}
 \le\Delta\le2r.
\]
Consequently $A(r)\le B(r)$, and both endpoint vectors meet the
claimed radius bounds, proving their extremality.
\end{proof}

For example, a zero-budget edge between scores $0,1$ and an isolated
score $0.2$ gives midpoint repair $(0.5,0.5,0.2)$. Computing
$U(h)=1-L(1-h)$ requires a second multi-source Dijkstra search;
the asymptotic time and storage remain unchanged. For a fixed graph,
the selector also obeys $M(1-h)=1-M(h)$ and
$\|M(h)-M(g)\|_\infty\le\|h-g\|_\infty$: each envelope is
nonexpansive in its input scores.

\paragraph{Secondary movement criterion.}
For positive weights $w_u$ summing to one, the program
\begin{equation}\label{eq:secondary_repair_v6}
 z^{\rm sq}
 =\arg\min_{z\in\mathcal C_{r_G(h)}}
       \sum_u w_u(z_u-h_u)^2
\end{equation}
selects the unique repair with least weighted squared movement among
those attaining the minimum uniform radius. Its feasible set is a
nonempty compact polytope and its
objective is strictly convex, giving a convex quadratic program. It leaves every
feasible component unchanged because keeping its input scores has
zero cost and imposes no constraints on other components. If
$\Pr(I=u)=w_u$ and $h_u=\Pr(Y=1\mid I=u)$, the objective equals the
increase in population Brier loss, so this selection minimizes that increase within
$\mathcal C_{r_G(h)}$. The secondary objective
can select a different optimal-radius repair from the midpoint.

\section{Joint adaptation under source retention and target uncertainty}
\label{app:joint_adaptation}

Fix a source graph $G_s$ and finitely many target graphs
$G_1,\ldots,G_p$, $p\ge1$, on the same $n\ge2$ profiles. All graph
budgets are finite and nonnegative. Let $\cF_s,\cF_j\subseteq[0,1]^n$
be their feasible classes and let $\rho_s,\rho_j$ be their capped
shortest-path pseudometrics. The graph $H$ is their union, with the
smallest available budget on each edge. Thus
$\cF_H=\cF_s\cap\bigcap_{j=1}^p\cF_j$.
For $h\in\cF_s$, use $R(h)=\max_jr_j(h)$ and $Q(h)=r_H(h)$ as in
\eqref{eq:joint_radii}.

\paragraph{Compatible paths and target runs.}
Give the source label $0$, and let $A_e\subseteq\{0,1,\ldots,p\}$
contain the standards attaining the union budget $c_e$ on edge $e$.
A compatible path assignment chooses a label in $A_e$ for every edge.
A target run is a maximal consecutive block with one fixed nonzero label.
Consecutive different target labels start different runs; a source edge
separates runs and contributes no run itself. For connected distinct
$u,v$, $k(u,v)$ is the minimum target-run count over compatible
assignments of simple shortest $H$-paths. Set $k(u,u)=0$ and
$K=\max_{d_H(u,v)<1}k(u,v)$ as in~\eqref{eq:joint_factor}.
Diagonal pairs make this maximum nonempty. Nonnegative budgets and finite
path and assignment sets ensure attainment, including with zero budgets.

\begin{proof}[Upper bound in Theorem~\ref{thm:joint_amplification}]
For a target-$j$ run from $x$ to $y$, the triangle inequality applied to an optimal target repair gives
\[
 |h_x-h_y|\le\sum_{e\text{ in the run}}c_e+2r_j(h)
 \le\sum_{e\text{ in the run}}c_e+2R(h).
\]
Source-labelled edges satisfy their budgets directly. Summing along a
compatibly labelled shortest union path attaining $k(u,v)$ gives
$|h_u-h_v|\le d_H(u,v)+2k(u,v)R(h)$.
Only pairs with $d_H(u,v)<1$ can contribute a positive score deficit.
The exact repair formula proves $Q(h)\le KR(h)$; set inclusion and
constant scores prove $R(h)\le Q(h)$ and $Q(h)\le1/2$.
A simple path has at most $n-1$ edges, proving $K\le n-1$.
With one target, $k$ target runs require at least $k-1$ intervening source
edges, so $2k-1\le n-1$ and $K\le\lfloor n/2\rfloor$.
\end{proof}

\paragraph{Exact feasible input class.}
Use $S$, $U$, and $w_a$ from Section~\ref{sec:joint_adaptation}.
The exact target-repair formula gives
$R(h)\le a$ if and only if
$|h_u-h_v|\le\rho_j(u,v)+2a$ for every $j,u,v$.
Together with source feasibility, these are exactly the $w_a$ constraints.
Consequently their feasible class is
\[
 \cF_a=\{h\in\cF_s:R(h)\le a\}.
\]
All direct budgets are at most one; hence their shortest-path closure
$\rho_a$ is at most one without another range cap. At $a=0$,
$\cF_a=\cF_H$ and $\rho_a=\rho_H$.
The directed comparison~\eqref{eq:directed_transfer_main_v10} proves
\eqref{eq:joint_cost_curve}. Since $\rho_a\ge\rho_H$, its differences
are nonnegative. For a maximizing pair $(u,v)$, the distance profile
$h_w=\rho_a(u,w)$ is in $\cF_a$ and has a deficit
$\rho_a(u,v)-\rho_H(u,v)$ on that pair. It therefore attains $C(a)$.

\paragraph{Exact min-plus representation.}
For min-plus multiplication,
$(A\otimes B)_{uv}=\min_w(A_{uw}+B_{wv})$, define
\[
 B_0=S,\qquad B_{m+1}=B_m\otimes U\otimes S,
 \quad 0\le m<n-1.
\]
Although each $\rho_j$ is a pseudometric, $U$ need not be one.
$B_m(u,v)$ is the least base length of a walk with $m$ target-closure
edges, each chosen from an arbitrary target, and source-closure paths
before, between, and after them. Empty source segments are permitted.
The extra budget $2a$ is charged once per target-closure edge. Therefore
\begin{equation}\label{eq:joint_full_envelope_app}
 \rho_a(u,v)=\min_{0\le m\le n-1}\{B_m(u,v)+2ma\}.
\end{equation}
Indeed, a simple shortest $w_a$ path has at most $n-1$ edges. Choose a
source or minimizing target label on each edge and group the source
segments. Conversely every walk on the right is a valid walk in the
complete budget graph. Source paths can be collapsed because $S$ is a
pseudometric; no triangle inequality for $U$ is used.

For a pair with $d=\rho_H(u,v)<1$,
\begin{equation}\label{eq:joint_first_attainment_app}
 k(u,v)=\min\{m:B_m(u,v)=d\}.
\end{equation}
To prove one direction, collapse every target run of a shortest compatible
union path to an edge in its target closure, and collapse its source
segments. The resulting base length is at most $d$. It cannot be smaller,
since $\rho_H\le S,\rho_j$ and $\rho_H$ is a pseudometric.
For the other direction, take a walk attaining $B_m(u,v)=d<1$ and choose
a target realizing each $U$ edge. No closure edge is capped at one, so
each expands to a finite shortest path in its own graph. Every expanded
edge must attain its union budget: a strict decrease would yield an
$H$-walk shorter than $d$. Every repeated-vertex cycle has zero length.
Deleting the corresponding contiguous sequence of edges cannot increase the number of target
runs: the two surviving boundary runs either remain separate or merge.
Deleting all such cycles gives a compatible simple shortest $H$-path
with at most $m$ target runs. The argument applies to both zero budgets and tied
minimum-budget labels.

\paragraph{Local attainment and the initial linear interval.}
For a pair attaining $k(u,v)=K>0$, define
\begin{equation}\label{eq:joint_initial_interval_app}
 a_{uv}=\min_{0\le m<K}
 \frac{B_m(u,v)-\rho_H(u,v)}{2(K-m)}.
\end{equation}
Every numerator is positive by~\eqref{eq:joint_first_attainment_app}, and
the $m=0$ term gives
$0<a_{uv}\le(1-\rho_H(u,v))/(2K)$.
Since $B_K(u,v)=\rho_H(u,v)$ and every $B_m(u,v)\ge\rho_H(u,v)$,
\eqref{eq:joint_full_envelope_app} gives
$\rho_a(u,v)=\rho_H(u,v)+2Ka$ on $[0,a_{uv}]$.
Thus $C(a)=Ka$ there. The profile $h_w=\rho_a(u,w)$ has
$R(h)\le a$ and $Q(h)\ge Ka$. The established inequality
$Q(h)\le KR(h)\le Ka$ forces $R(h)=a$ and $Q(h)=Ka$ for $a>0$.
No smaller coefficient works uniformly over the fixed source class.
If $K=0$, the same inequality gives $Q(h)=R(h)=0$ for every source-feasible
input, proving $\cF_s\subseteq\bigcap_j\cF_j$ and completing the
coefficient statements of Theorem~\ref{thm:joint_amplification}.

\paragraph{Approximate source feasibility.}
For arbitrary $h\in[0,1]^n$, let $s=r_s(h)$ and let $h^0$ be an optimal
source repair, so $h^0\in\cF_s$ and $\|h^0-h\|_\infty=s$.
Distance to a fixed nonempty feasible set is $1$-Lipschitz, giving
$r_j(h^0)\le r_j(h)+s$ for every target. Hence
\[
 r_H(h)\le s+r_H(h^0)
 \le s+K\max_jr_j(h^0)
 \le (K+1)s+K\max_jr_j(h).
\]
The constant-score repair supplies the $1/2$ cap in
\eqref{eq:joint_inexact_source}. At both stages, the distance envelopes construct optimal
repairs. When $K=0$, inclusion of the source class in
every target class gives $\cF_H=\cF_s$, so $r_H(h)=r_s(h)$.

The largest initial interval on which $C(a)=Ka$ is exactly
$[0,a_{\rm lin}]$, where
\[
 a_{\rm lin}=\max_{(u,v):\,k(u,v)=K,\,\rho_H(u,v)<1}a_{uv}.
\]
For $a>0$, a pair with $k<K$ has deficit at most $2ka<2Ka$;
a pair with $k=K$ has deficit $2Ka$ exactly until $a_{uv}$.

\paragraph{Cost-curve structure and computational complexity.}
For a pair with $d=\rho_H(u,v)<1$ and $k=k(u,v)$, $B_k=d$ and
$B_m\ge d$ for all $m$. Every term with $m>k$ is therefore dominated by
$m=k$ in~\eqref{eq:joint_full_envelope_app} for all $a\ge0$.
If $\rho_H(u,v)=1$, then $\rho_a=S=1$, so that pair contributes zero.
Thus only $B_0,\ldots,B_K$ are needed, proving~\eqref{eq:joint_cost_envelope}.
Each pairwise contribution to $C$ is the lower envelope of lines with
slopes in $\{0,1,\ldots,k\}$. It is nondecreasing and $K$-Lipschitz.
A finite maximum preserves those properties and piecewise linearity;
every open linear segment of the resulting envelope has one of the
original slopes.

Let
\[
 D=D(s\!\to\!H)=\tfrac12\max_{u,v}\{S(u,v)-\rho_H(u,v)\}.
\]
The source term $B_0=S$ bounds $C$ by $D$ and becomes minimal for every
pair at a sufficiently large $a$. Hence $C(a)\le\min\{Ka,D\}$ and
eventually $C(a)=D$.
If $K=1$, each positive pairwise contribution is
$\min\{a,(S(u,v)-\rho_H(u,v))/2\}$; maximizing gives
$C(a)=\min\{a,D\}$. This proves all claims in
Theorem~\ref{thm:joint_cost_curve}.

Once $K$ and the standard closures are available, the recurrence takes
$O(Kn^3+n^2)$ arithmetic operations and $O((K+1)n^2)$ storage using
ordinary min-plus products. For one specified $a$, dense all-pairs
shortest paths instead evaluate $C(a)$ in $O(n^3)$ time and $O(n^2)$
storage. These bounds describe an exact finite representation; evaluating
or listing its envelopes is a separate operation.

\paragraph{Exact saturation threshold.}
For $D>0$, define
$\mathcal V=\{(u,v):S(u,v)-\rho_H(u,v)=2D\}$.
The earliest budget at which $C(a)=D$ is
\begin{equation}\label{eq:joint_saturation_app}
 a_{\rm sat}=\min_{(u,v)\in\mathcal V}\ 
 \max_{1\le m\le K}\frac{[S(u,v)-B_m(u,v)]_+}{2m}.
\end{equation}
For a maximizing pair, the cost attains $D$ exactly when
$\rho_a(u,v)=S(u,v)$, which by~\eqref{eq:joint_cost_envelope} is the
displayed set of linear inequalities. Set $a_{\rm sat}=0$ when $D=0$.
The upper bound
\[
 a_{\rm sat}\le
 \max_j D(s\!\to\!j)
 =\tfrac12\max_{u,v}\{S(u,v)-U(u,v)\}
\]
follows because above the latter budget all source-feasible scores are
admissible inputs.

\paragraph{Nonconcavity of the cost curve.}
Take two disconnected components, one on two vertices and one on four.
On the two-vertex component, the source edge budget is $.8$ and the target
edge budget is $.2$; its cost curve is $C_1(a)=\min\{a,.3\}$.
On the four-vertex component, the source is a path with all three budgets
$.2$; the target contains its first and third edges, both with budget
$.1$. Its curve is $C_2(a)=\min\{2a,.1\}$: the target-only constraint
limits each matching-edge deficit to $2a$, source feasibility limits
each to $.1$, and nondecreasing scores attain both bounds simultaneously.
In detail, put $b=\min\{2a,.1\}$ and use consecutive score gaps
$(.1+b,.2,.1+b)$. The endpoint score is $.4+2b\le.6$; the source
constraints hold, target-only cost is $b/2\le a$, and union repair costs
$b$. The general factor and directed-class bound prove its optimality.
Disconnected components optimize independently, giving
\[
 C(a)=\max\{\min(a,.3),\min(2a,.1)\}.
\]
Its successive slopes are $2,0,1,0$, with breakpoints $.05,.1,.3$.
Hence $C$ need not be concave and can differ from $\min\{Ka,D\}$.

\paragraph{Lexicographic computation of $K$.}
For each vertex $v$, create a hub $H_v$, a source state $(v,0)$, and a
state $(v,j)$ for each target label $j$ appearing on an incident union
edge. Include zero-cost arcs from each state at $v$ to $H_v$, an arc
$H_v\to(v,0)$ of cost $(0,0)$, and arcs $H_v\to(v,j)$ of cost $(0,1)$
for $j>0$. For every compatible label $j\in A_e$ on edge $e=(v,w)$,
include both traversal arcs $(v,j)\leftrightarrow(w,j)$ of cost $(c_e,0)$.
Costs are ordered lexicographically: length first, then target-run count.
Start at $(u,0)$. Remaining in a target state continues its run without increasing the run count;
entering a target through a hub starts a new run and increments the run count. The smallest
terminal cost at $v$ is $(d_H(u,v),k(u,v))$.

Every compatible path has the corresponding expanded walk. Conversely,
an expanded walk induces a compatible union walk whose true run count is
no larger than the number of charged switches; switches with no traversal are
unnecessary. A lexicographic optimum therefore agrees with the path
statistic. Nonnegative costs permit Dijkstra's algorithm, and zero-length cycle
deletion gives the simple-path convention established above.
With $L=\sum_{e\in E_H}|A_e|$, this graph has $O(n+L)$ states and arcs.
Running the search from every starting vertex computes $K$ in
$O(n(n+L)\log(n+L))$ time and $O(n+L)$ working space, assuming exact
comparisons of budget sums. All tied labels must be retained.

\paragraph{Sharp special cases.}
With one target, the alternating source/target construction in
Appendix~\ref{app:source_preservation_v11} attains
$K=\lfloor n/2\rfloor$ with positive budgets.
With an unconstrained source and two alternating matching targets of
budget $\gamma/(n-1)$, $0\le\gamma<1$, their union path has $K=n-1$.
The endpoint path has total budget $\gamma<1$ and alternates target
labels at every edge. The fixed two-dimensional metric construction in
Proposition~\ref{prop:alternating_geometry_v9} realizes these targets.
Thus the joint comparison recovers both sharp factors and bounds their
combined effect by $n-1$.

\paragraph{Refinement of the target family.}
For a fixed predictor, discarding plausible targets can only decrease the
common-repair cost. The relative factor $K$ need not decrease. With an
unconstrained source on three vertices, let targets have zero-budget
edges $\{(1,2)\}$, $\{(2,3)\}$, and $\{(1,2),(2,3)\}$ respectively.
With all three, the third realizes the common graph and $K=1$.
Removing it leaves the common graph unchanged but gives $K=2$.
Elicitation therefore cannot increase the absolute common movement
certificate, even when its ratio to revealed repair becomes larger.

\subsection{Relation to relative Hoffman constants}
\label{app:joint_hoffman}

The joint amplification theorem gives a relative error bound with the
right-hand side held fixed. The path characterization provides an explicit
value for this bound, complementing general existence and variational
characterizations of Hoffman constants.
Let $P=\cF_s$. For every target $j$ and ordered pair $u\ne v$, put
\[
 (Az)_{j,u,v}=z_u-z_v,\qquad b_{j,u,v}=\rho_j(u,v).
\]
These are the implied inequalities obtained by closing each target
separately. For this fixed $b$, define
\[
 \mathsf H_b(A\mid P)=\inf\bigl\{c\ge0:
 \operatorname{dist}_\infty(h,P\cap\{z:Az\le b\})
 \le c\|(Ah-b)_+\|_\infty\quad\text{for all }h\in P\bigr\}.
\]

\begin{proposition}[Fixed-budget relative error constant]
\label{prop:joint_hoffman}
For the closed target system above,
\[
 P\cap\{z:Az\le b\}=\cF_H,\qquad
 \|(Ah-b)_+\|_\infty=2R(h),\qquad
 \mathsf H_b(A\mid P)=K/2.
\]
The final identity includes $K=0$, when every $h\in P$ is already
feasible for all targets.
\end{proposition}
\begin{proof}
The first equality follows from the path-closure description of each
feasible class. For each target, the maximum ordered-pair residual is
$2r_j(h)$ by~\eqref{eq:repair_main}; maximizing over targets gives the
second equality. The smallest coefficient relating $Q(h)$ to $R(h)$
on $P$ is $K$ by Theorem~\ref{thm:joint_amplification}, which proves
the last identity, including its zero case.
\end{proof}

The relative constant $\mathsf H(A\mid P)$ of
\citet{pena2021hoffman} instead requires one bound uniformly over all
right-hand sides $b'$ for which $P\cap\{z:Az\le b'\}$ is nonempty.
Their characterizations use row subsets and tangent cones of $P$.
Consequently, $\mathsf H_b(A\mid P)\le\mathsf H(A\mid P)$, and equality
need not hold. For example, take $P=[0,1]^2$, one target requiring
$z_1=z_2$, and the two rows $(1,-1)$ and $(-1,1)$ of $A$.
For $b=(0,0)$, $K=1$ and
$\mathsf H_b(A\mid P)=1/2$. The uniform bound must also cover
$b'=(1,-1)$, whose feasible set in $P$ is $\{(1,0)\}$.
At $h=(0,0)$, both the distance and residual norm equal one, so
$\mathsf H(A\mid P)\ge1$.

Closing each target separately identifies the intended residual:
raw edge residuals need not equal twice the target repair cost, whereas
closing the entire union gives $2Q(h)$ and removes the target-specific
comparison. The target-run theorem computes the fixed-$b$ constant for
these separately closed systems by counting target runs and constructs
score vectors attaining the bound. The cost curve strengthens this
constant bound further:
$C(a)$ is the exact maximum distance to feasibility on $P$ subject to
$\|(Ah-b)_+\|_\infty\le2a$. Its complete envelope describes behavior
at finite residual levels, including plateaus and later increases.

\section{Source-retention cost}
\label{app:source_preservation_v11}

\begin{proof}[Proof of Corollary~\ref{cor:source_runs_main_v16}]
Theorem~\ref{thm:joint_amplification} with one target gives the exact
coefficient $K$, and Theorem~\ref{thm:joint_cost_curve} shows that this factor is attained for sufficiently small budgets and
gives the full cost curve. A simple union path with $k$ target
runs needs at least $k-1$ separating source edges. Thus
$2k-1\le n-1$ and $K\le\lfloor n/2\rfloor$.
If every union budget is at least $c>0$, a path contributing to $K$
has $m$ edges with $mc<1$, hence
$m\le\lceil1/c\rceil-1$. Its target-run count is at most
$\lceil m/2\rceil\le\lfloor\lceil1/c\rceil/2\rfloor$.
The construction below proves sharpness with positive tolerances and
fixed two-dimensional Mahalanobis metrics.
\end{proof}

\paragraph{Sharpness with positive edge budgets.}
Let $a=1/(n-1)$, $b=\gamma a$ with $0<\gamma<1$, and
$h_i=(i-1)a$. The source graph is the full path
$1$--$2$--$\cdots$--$n$, with budget $a$ on every edge. The target
contains its $q=\lfloor n/2\rfloor$ odd-indexed edges, with budget $b$.
The source scores are feasible, and every disjoint target edge has
deficit $a-b$. On the union path, an interval's deficit is its number
of odd edges times $a-b$; the endpoint pair contains all $q$. Therefore
\begin{equation}\label{eq:source_preservation_sharp_v11}
 r_t(h)=\frac{a-b}{2},\qquad
 r_{s\wedge t}(h)=\frac{q(a-b)}{2}.
\end{equation}
This attains the claimed factor for every $n\ge2$.

For the metric realization, set $x_1=0$ and let $x_{i+1}-x_i$ be $e_1$
for odd $i$ and $e_2$ for even $i$. Use threshold one and
\[
 M_s=\tfrac9{10}I_2,
 \qquad M_t=\operatorname{diag}(2/3,2).
\]
Under $M_s$, adjacent squared distances are $9/10<1$, while every
nonadjacent squared distance is at least $9/5>1$. Under $M_t$, odd
and even adjacent squared distances are $2/3$ and $2$; every nonadjacent
squared distance is at least $4/3>1$. Thus the full pair domain gives
exactly the source path and target matching. The matrices and activation
margins are fixed independently of $n$; the positive tolerances $a,b$
are of order $1/n$, and the profile diameter grows with $n$.

\subsection{Equal repair costs under source retention}
\label{app:source_runs_v16}

If the union is a tree with a compatible assignment whose target-labelled
edges form a connected subtree, every simple path crosses those edges
in at most one run. Then $K\le1$, independently of pool size, so source
retention adds no optimal uniform movement. The repairs themselves can
still differ, as the following positive-tolerance example shows.

Take four profiles in $[0,1]^2$,
\[
 x_1=(0,0),\quad x_2=(1/2,0),\quad
 x_3=(1/2,1/2),\quad x_4=(1/2,1),
\]
and fixed positive-definite matrices
\[
 M_s=\operatorname{diag}(8/3,8),\qquad
 M_t=\operatorname{diag}(8,8/3).
\]
At threshold one, the source constrains only $(1,2)$ and the target
constrains $(2,3),(3,4)$. Their active squared distances are $2/3$;
the other adjacent squared distances are two and all nonadjacent distances
exceed one. Give the source edge budget $1/4$ and each target edge
budget $1/8$. The union labels are source--target--target, so $K=1$
instead of the worst-case factor $\lfloor4/2\rfloor=2$.

For the source-feasible scores $h=(0,1/4,3/4,1)$,
\[
 \begin{aligned}
 r_t(h)&=r_{s\wedge t}(h)=1/4,\\
 z_t&=(0,1/2,5/8,3/4),\\
 z_{s\wedge t}&=(1/4,1/2,5/8,3/4).
 \end{aligned}
\]
The target-only midpoint $z_t$ violates the source edge; the second
repair satisfies both standards with the same maximum coordinate change. Both radii are
certified by a deficit of $1/2$, using pair $(2,4)$ for the target
and $(1,4)$ for the union. Retaining the source changes an additional
coordinate even though the optimal uniform costs agree.

\section{Repair under specification uncertainty}
\label{app:uncertainty}

\subsection{Repair-budget certificates under elicitation uncertainty}
\label{app:uncertain_repair}

Fix a pool $V=\{1,\ldots,n\}$ and a declared set $\mathcal P$ of candidate
unordered pairs. Both the partition and the permitted comparisons are
fixed. For each $e=(u,v)\in\mathcal P$, the unknown threshold specification
imposes $|z_u-z_v|\le\epsilon_e^\star$ when $d_e^\star\le1$.
The tolerance may be shared by every pair in a community. Suppose the supplied
bands satisfy the simultaneous event
\begin{equation}\label{eq:band_event_v5}
 \mathcal E=\{\ell_e\le d_e^\star\le u_e,\quad
               \epsilon_e^-\le\epsilon_e^\star\le\epsilon_e^+
               \text{ for all }e\in\mathcal P\},
 \qquad \Pr(\mathcal E)\ge1-\delta,
\end{equation}
where $0\le \ell_e\le u_e$ and
$0\le\epsilon_e^-\le\epsilon_e^+<\infty$. Known tolerances give degenerate
tolerance intervals.

Construct the safe graph $G^-$ and relaxed graph $G^+$ on $V$:
\begin{align*}
 E^-&=\{e\in\mathcal P:\ell_e\le1\},
 &c_e^-&=\epsilon_e^-,\\
 E^+&=\{e\in\mathcal P:u_e\le1\},
 &c_e^+&=\epsilon_e^+.
\end{align*}
Thus the safe graph constrains every possibly required pair at the smaller
budget; the relaxed graph constrains only certainly required pairs at the
larger budget. Write $\mathcal F^-,\mathcal F_\star,
\mathcal F^+$ for the $[0,1]$-valued score classes satisfying the safe,
true, and relaxed specifications, respectively. Let $r^-(h),r_\star(h),r^+(h)$
be their respective minimum $\ell_\infty$ repair radii.

\begin{theorem}[Finite-pool repair certificate under specification uncertainty]
\label{thm:uncertain_repair}
On $\mathcal E$, simultaneously for every $h\in[0,1]^n$,
\begin{equation}\label{eq:class_sandwich_v5}
 \mathcal F^-\subseteq\mathcal F_\star
 \subseteq\mathcal F^+,\qquad
 r^+(h)\le r_\star(h)\le r^-(h).
\end{equation}
Both endpoints are exactly computable from the corresponding capped
shortest-path pseudometrics using
\[
 r_G(h)=\frac12\max_{u,v\in V}
 [|h_u-h_v|-\rho_G(u,v)]_+.
\]
In particular, for any movement budget $t\ge0$:
\begin{enumerate}
 \item If $r^-(h)\le t$, the distance-envelope construction on
 the safe graph returns a score satisfying the true specification with
 $\|z-h\|_\infty\le t$.
 \item If $r^+(h)>t$, no score satisfying the true
 specification has $\|z-h\|_\infty\le t$.
 \item Otherwise these bounds do not decide whether such a repair exists.
\end{enumerate}
The joint confidence event makes these statements valid even when $h$ or $t$
are selected after observing the bands.
\end{theorem}

\begin{proof}
Every true edge has $d_e^\star\le1$, so $\ell_e\le1$ and it is included in
the safe graph. A safe score has gap at most
$\epsilon_e^-\le\epsilon_e^\star$ on that edge, proving the first
inclusion. Conversely, every relaxed edge has
$d_e^\star\le u_e\le1$ and hence is a true edge. A score satisfying the true specification
has gap at most $\epsilon_e^\star\le\epsilon_e^+$ there, proving the
second inclusion. Taking distances from $h$ reverses the inclusion
order and yields the radius bounds. Proposition~\ref{prop:exact_pool_repair_v3}
computes each endpoint and constructs a safe minimizer. The two decisions
follow directly. All arguments hold for every $h,t$ on the same event.
\end{proof}

\paragraph{Exact endpoints and conservative bounds.}
Coordinatewise bands can combine distances from different members of a
coupled metric family. Although each endpoint is the exact repair radius
for its constructed graph, optimizing over the full coupled confidence
set may tighten the interval while preserving a safe repair.

\begin{corollary}[Directional class comparison under uncertainty]
\label{cor:uncertain_transfer_v5}
Suppose source and target bands hold jointly. Let superscripts
$-$ and $+$ denote their safe and relaxed graphs. For
$D(\mathcal F_s,\mathcal F_t)
=\sup_{h\in\mathcal F_s}\inf_{z\in\mathcal F_t}\|h-z\|_\infty$,
\begin{align*}
 \frac12\max_{u,v}
 [\rho_s^-(u,v)-\rho_t^+(u,v)]_+
 &\le D(\mathcal F_s^\star,\mathcal F_t^\star)\\
 &\le\frac12\max_{u,v}
 [\rho_s^+(u,v)-\rho_t^-(u,v)]_+.
\end{align*}
\end{corollary}
\begin{proof}
The directed distance increases when the source class is enlarged and
decreases when the target class is enlarged. Apply the class inclusions
in Theorem~\ref{thm:uncertain_repair}, then the exact directed
graph identity to the two extreme class pairs.
\end{proof}

\subsection{Sensitivity to edge budgets and neighborhood membership}
\label{app:repair_stability_v5}

\begin{proposition}[Sharp fixed-topology sensitivity]
\label{prop:repair_stability}
Let $G$ and $G'$ have the same $n$ vertices and edge set, with nonnegative
edge lengths satisfying $\max_e|c_e-c'_e|\le\eta$. Then for every
$h\in[0,1]^n$,
\begin{equation}\label{eq:repair_stability_v5}
 |r_G(h)-r_{G'}(h)|
 \le\frac12\|\rho_G-\rho_{G'}\|_\infty
 \le\frac12\min\{1,(n-1)\eta\}.
\end{equation}
The factor $1/2$ and the dependence on path length cannot in general be
improved.
\end{proposition}
\begin{proof}
The exact radius formula, the Lipschitz property of the positive-part map, and
$|\max f-\max g|\le\max|f-g|$ prove the first bound. Within a connected
component, a shortest path can be taken simple and has at most $n-1$
edges. Evaluating a shortest path for one graph under the other graph's
edge lengths, in both directions, gives
$|d_G(u,v)-d_{G'}(u,v)|\le(n-1)\eta$. The components are the same in both
graphs; disconnected pairs have capped distance one in both. Range
capping proves the second bound.

For sharpness, take an $n$-vertex path, $n\ge2$, scores
$h_i=(i-1)/(n-1)$, and constant edge lengths $c_e=a$ and
$c'_e=a+\eta$, where $a\ge0$ and $(n-1)(a+\eta)\le1$. The endpoint pair
maximizes the deficit, and the two radii are
$[1-(n-1)a]/2$ and $[1-(n-1)(a+\eta)]/2$. Their difference is
$(n-1)\eta/2$. With $a=0$ and $\eta\ge1/(n-1)$ the radii are $1/2$ and
zero, attaining the range cap as well.
\end{proof}

If every supplied distance interval lies wholly on one side of the
threshold (either $u_e\le1$ or $\ell_e>1$), the safe and relaxed graphs have
the same topology. A tolerance-band width at most $\eta$ then bounds
the radius interval width by $\tfrac12\min\{1,(n-1)\eta\}$.
Uniform distance accuracy together with a margin away from one ensures
that the two graphs have the same edges. When a band crosses the
threshold, the repair-radius interval accounts for the pair's uncertain
membership.

\subsection{Predictive utility under a certified movement budget}
\label{app:repair_utility_v5}

\begin{proposition}[Predictive utility bounds under a uniform movement budget]
\label{prop:repair_utility}
Let $h,z\in[0,1]^n$ satisfy $\|z-h\|_\infty\le r\le1$, and let
$p_i\ge0$ sum to one. Then:
\begin{enumerate}
 \item A decision thresholded at $1/2$ can change only on indices with
 $|h_i-1/2|\le r$. For any fixed binary labels $y_i$, the absolute change
 in weighted classification accuracy is at most
 $\sum_i p_i\mathbf1\{|h_i-1/2|\le r\}$.
 \item For any binary labels $y_i$,
 \[
 \left|\sum_i p_i\{(z_i-y_i)^2-(h_i-y_i)^2\}\right|
 \le2r-r^2.
 \]
 The pointwise constant $2r-r^2$ is sharp.
 \item If $I$ has distribution $p$ and the source scores are the true
 conditional event probabilities, $h_i=\mathbb E[Y\mid I=i]$, then
 the population Brier increase of the fixed repair is exactly
 \[
 \mathbb E[(z_I-Y)^2]-\mathbb E[(h_I-Y)^2]
 =\sum_i p_i(z_i-h_i)^2\le r^2.
 \]
\end{enumerate}
\end{proposition}
\begin{proof}
The first assertion follows because two scores on different sides of
$1/2$ differ by at least the source score's distance from that threshold;
changing correctness requires a decision change. For the second, use
symmetry in $y$ and in $h,z$ to consider $y=0$ and $z=h+t\ge h$, where
$0\le t\le r$ and $h\le1-t$. Then
$z^2-h^2=2ht+t^2\le2t-t^2\le2r-r^2$. Equality holds at
$h=1-r,z=1,y=0$. Weighted averaging gives the displayed claim.
For the final assertion, expand the squared losses. Conditional on $I$,
the cross term $2(z_I-h_I)(h_I-Y)$ has expectation zero. The remaining
term is the stated weighted squared movement.
\end{proof}

\paragraph{Conditional probabilities and calibration.}
The first two conclusions apply to arbitrary scores, including uncalibrated
predictions. For profile-dependent repairs, the Brier identity uses the
conditional-probability assumption $h_I=\mathbb E[Y\mid I]$ to eliminate
the cross term. Ordinary calibration $\mathbb E[Y\mid h_I]=h_I$ suffices
when the repair depends only on the source score. These population statements
concern fixed scores and a declared label distribution, or fresh labels after
conditioning on independent training data.

\section{Common and revealed repair costs}
\label{app:robust_common_v6}

Theorems~\ref{thm:common_main_v6} and~\ref{thm:price_main_v6}
establish the common graph and the universal factor. This appendix gives
the sharp metric construction and proves the bounds under a positive
tolerance floor and a bounded profile domain.
For a singleton pool both repair costs are zero; below $n\ge2$.

\subsection{Sharpness with a fixed uncertainty ball and a fixed activation margin}
\label{app:alternating_geometry_v9}

The construction uses only matching and empty scenario graphs. Their union
forms a path, so a repair selected before revelation must satisfy
constraints that propagate across the pool.

\begin{proposition}[Two alternating matchings in a fixed two-dimensional ball]
\label{prop:alternating_geometry_v9}
For every $n\ge2$, there are $n$ distinct profiles in $\mathbb R^2$
such that, for the same full-dimensional convex uncertainty set
\begin{equation}\label{eq:alternating_ball_v9}
 \mathcal C=\{M=M^\top:\|M-2I_2\|_F\le\tfrac75\},
\end{equation}
every threshold graph on the full pair domain is either empty or one
of the two alternating matchings of the path $1,\ldots,n$.
The matrices in $\mathcal C$ satisfy
$(3/5)I_2\preceq M\preceq(17/5)I_2$.
Each nonempty matching occurs at an interior matrix where all its
edges have squared distance $2/3$, uniformly in $n$.
The union graph is the full path.

For source scores $h_i=(i-1)/(n-1)$ and common known tolerance
$\epsilon=\gamma/(n-1)$, $0\le\gamma<1$,
\begin{equation}\label{eq:alternating_radii_v9}
 R_{\rm rev}=\frac{1-\gamma}{2(n-1)},\qquad
 R_{\rm com}=\frac{1-\gamma}{2}.
\end{equation}
The same equalities hold when the uncertainty set contains only the
two interior matrices described below.
\end{proposition}

\begin{proof}
Let $e_1,e_2$ be the coordinate vectors, and set $x_1=0$ and
\[
 x_{i+1}-x_i=v_i=
 \begin{cases}
 e_1,&i\text{ odd},\\
 e_2,&i\text{ even}.
 \end{cases}
\]
The coordinate sum of $x_i$ is $i-1$, so the profiles are distinct.
The ball is compact and has nonempty interior in the three-dimensional
space of symmetric matrices. Its spectral bounds follow from
$\|M-2I_2\|_{\rm op}\le\|M-2I_2\|_F\le7/5$.

Every odd adjacent edge has squared distance $M_{11}$, and every even
adjacent edge has squared distance $M_{22}$. Thus an orientation is
active on all its adjacent edges precisely when its corresponding
diagonal entry is at most one. Both diagonal entries cannot be at most
one in the ball: this would require
\[
 \|M-2I_2\|_F^2
 =(M_{11}-2)^2+(M_{22}-2)^2+2M_{12}^2
 \ge2>\tfrac{49}{25}.
\]
Consequently every scenario activates at most one of the two
orientations. The interior matrices
\[
 M^{(1)}=\operatorname{diag}(2/3,2),\qquad
 M^{(2)}=\operatorname{diag}(2,2/3)
\]
have Frobenius displacement $4/3<7/5$ and realize the odd and even
matchings, respectively. Their active squared distance is $2/3$,
so their squared-distance activation margin is $1/3$; the inactive
orientation has margin $1$ above threshold. The center realizes the
empty graph. When $n=2$, the even matching itself is empty.

For nonadjacent indices $u<v$, put $\ell=v-u\ge2$. The displacement
$x_v-x_u$ has nonnegative integer coordinates summing to $\ell$.
It follows that
\[
 \|x_v-x_u\|^2\ge\ell^2/2\ge2,
 \qquad
 (x_v-x_u)^\top M(x_v-x_u)\ge(3/5)\,2=6/5>1.
\]
Hence there are no nonadjacent edges in any scenario. This also proves
the claim for the full pair domain rather than a preselected path.

On a matching, the constrained edges have disjoint endpoints. For any
scores in $[0,1]^n$ and common edge tolerance $\epsilon$, its repair
radius is
\[
 \frac12\max_{(u,v)\text{ in the matching}}
       [|h_u-h_v|-\epsilon]_+,
\]
with zero for an empty matching: each violating pair can be moved
symmetrically toward its midpoint, without interacting with other
edges or leaving the score interval. For the displayed ramp, every
adjacent source gap is $1/(n-1)$ and some matching is nonempty.
This gives the first equality in~\eqref{eq:alternating_radii_v9}.

A common repair satisfies the full path. Its endpoint gap may not
exceed $(n-1)\epsilon=\gamma$, so reducing the original endpoint gap
of one requires maximum movement at least $(1-\gamma)/2$. The scores
\[
 z_i=\frac{1-\gamma}{2}+\gamma\frac{i-1}{n-1}
\]
meet every path constraint and attain that movement. This proves the
second equality. Both matrices suffice to realize the full
union path and the same maximal matching deficit, so restricting to
these two standards leaves both costs unchanged.
\end{proof}

\paragraph{Dependence on pool size.}
The feature dimension, spectral bounds, uncertainty ball, and strict
activation margin at these matrices are fixed independently of $n$.
There are at most three distinct target graphs. The domain diameter grows
with pool size:
\[
 \operatorname{diam}\{x_1,\ldots,x_n\}
 =\sqrt{\lceil(n-1)/2\rceil^2+\lfloor(n-1)/2\rfloor^2}
 =\Theta(n).
\]
Attaining the full factor with nonzero costs uses tolerance
$O(1/n)$, including zero; a fixed positive tolerance is subject to
Corollary~\ref{cor:positive_tolerance_v6}. The fixed activation margin
holds at the two specified interior matrices. Over the full ball,
matrices can approach the activation threshold arbitrarily closely.

\begin{corollary}[Sharp bounds with a positive tolerance floor]
\label{cor:positive_tolerance_v6}
Suppose every active edge budget in every scenario is at least a fixed
$c>0$, and put
\[
 K_c=\min\{n-1,\max\{0,\lceil1/c\rceil-1\}\}.
\]
If $K_c=0$, both repair costs are zero. Otherwise,
\begin{align}
 R_{\rm rev}\le R_{\rm com}
 &\le\min\{1/2,K_c R_{\rm rev}\},
 \label{eq:positive_tolerance_factor_v6}\\
 R_{\rm com}-R_{\rm rev}
 &\le\max_{1\le k\le K_c}
       \frac{(k-1)(1-kc)}{2k}.
 \label{eq:positive_tolerance_additive_v6}
\end{align}
For each fixed $n\ge2$ and $0<c<1$, the factor $K_c$ and the additive
bound are sharp even when every active edge has the same known budget
$c$ and uncertainty is the two-dimensional Mahalanobis ball of Proposition~\ref{prop:alternating_geometry_v9}.
\end{corollary}

\begin{proof}
The common graph has edge budgets $b_e\ge c$. A positive score deficit
can only occur on a pair with shortest-path length less than one.
If a simple shortest path for that pair has $k$ edges, then
$kc\le d_{G_\cap}(u,v)<1$, hence
$k\le\min\{n-1,\lceil1/c\rceil-1\}=K_c$.
Replacing $n-1$ by $K_c$ in the path argument of
Theorem~\ref{thm:price_main_v6} proves the factor bound.
If no such path exists, $R_{\rm com}=R_{\rm rev}=0$.

For the additive bound, if $R_{\rm com}>0$, choose a pair attaining
its maximum deficit and a simple shortest path with $k$ edges. The
same path argument gives $R_{\rm com}\le kR_{\rm rev}$, while
the score range and path budget give
$R_{\rm com}\le(1-kc)/2$. Therefore
\[
 R_{\rm com}-R_{\rm rev}
 \le\left(1-\frac1k\right)R_{\rm com}
 \le\frac{(k-1)(1-kc)}{2k}.
\]

For sharpness, use the profiles and matrix ball of
Proposition~\ref{prop:alternating_geometry_v9}, whose scenario graphs are the two alternating matchings or the empty graph, and whose union is the full path. For any integer
$1\le k\le K_c$, set every active budget to $c$ and take
$h_i=\min\{(i-1)/k,1\}$. The largest adjacent source gap is $1/k$,
so $R_{\rm rev}=(1/k-c)/2$. For a pair separated by $\ell$ path
edges its gap is at most $\min\{\ell/k,1\}$, so its positive deficit
is at most $[\min\{\ell/k,1\}-\ell c]_+$. This upper bound increases
up to $\ell=k$ and cannot increase beyond it. The pair
$(1,k+1)$ attains deficit $1-kc$, so
$R_{\rm com}=(1-kc)/2$. Taking $k=K_c$ attains the factor; choosing
an integer maximizing~\eqref{eq:positive_tolerance_additive_v6}
attains the additive bound.
\end{proof}

\subsection{Bounded profile domains}
\label{app:bounded_domain_v9}

\begin{proposition}[Commitment bound under bounded geometry]
\label{prop:bounded_domain_v9}
Suppose all candidate pairs are included, $x_1,\ldots,x_n\in\mathbb R^d$
have Euclidean diameter at most $D$, with $d\ge1$, and every metric
scenario satisfies $0\prec M\preceq BI$ for a fixed $B>0$. Let the activation threshold be one
and let every active edge have the same known tolerance $\epsilon\ge0$.
Put
\[
 P=\left\lfloor(1+2D\sqrt B)^d\right\rfloor,\qquad
 H=\min\{n-1,2P-1\}.
\]
For $n\ge2$,
\[
 R_{\rm rev}\le R_{\rm com}
 \le\min\{1/2,H R_{\rm rev}\}.
\]
If $\epsilon>0$, the factor can be further reduced to
$\min\{H,\max\{0,\lceil1/\epsilon\rceil-1\}\}$,
interpreting factor zero as both costs being zero.
In particular, at fixed $d,D,B$ the factor is bounded independently
of pool size, including when $\epsilon=0$. For $n=1$, both repair
costs are zero.
\end{proposition}

\begin{proof}
Every profile pair at Euclidean distance at most $r=1/\sqrt B$ has
squared Mahalanobis distance at most one for every scenario, and is
therefore an edge of the common graph. Consider a path
$u_0,\ldots,u_k$ with the fewest edges between its endpoints. It is simple
and has no edge between nonconsecutive vertices, since such an edge would shorten the path.
Consequently any two distinct vertices with even indices, and
likewise any two with odd indices, are at Euclidean distance greater
than $r$.

For either parity class, the open Euclidean balls of radius $r/2$
around its profiles are disjoint. All are contained in the ball of
radius $D+r/2$ about any one of the original profiles. Comparing
volumes bounds that class's size by
$(D+r/2)^d/(r/2)^d=(1+2D/r)^d$, hence by $P$.
Both parity classes together give $k+1\le2P$, and simplicity gives
$k\le n-1$. Therefore $k\le H$.

When $\epsilon>0$, paths with the fewest edges also minimize the total
common edge budget. When $\epsilon=0$, every path minimizes that budget,
so we may again select one with the fewest edges. The edge deficit bound
$|h_u-h_v|\le\epsilon+2R_{\rm rev}$ sums along this path to at most
$2H R_{\rm rev}$ beyond its allowed total gap. The exact graph repair
formula yields the displayed bound. Finally, a positive deficit on a
path with $k$ edges requires $k\epsilon<1$, yielding the additional
positive-tolerance cap as in Corollary~\ref{cor:positive_tolerance_v6}.
\end{proof}

\paragraph{Geometry and the common tolerance.}
The constant comes from a volume packing estimate. The full pair domain
ensures that nearby profiles supply shortcuts, and the common tolerance
ensures that these shortcuts preserve or reduce path cost. With omitted pairs or
heterogeneous budgets, these implications can fail. Under the stated
assumptions, the cap explains why the diameter grows in the sharpness
construction of Proposition~\ref{prop:alternating_geometry_v9}.

\section{Pairwise elicitation for movement-budget certification}
\label{app:direct_elicitation_v6}

Fix a finite pool, a candidate pair set $\mathcal P$ of size $m\ge1$,
known edge tolerances $\epsilon_e\ge0$, and source scores $h\in[0,1]^n$.
For each pair $e$, assume a stream of independent Bernoulli similarity
judgments with a fixed mean $p_e$. A sampling policy may adaptively reveal
the next unused response from any stream.
The target specification constrains pair $e$ exactly when $p_e\ge1/2$.
Under the logistic oracle, $p_e=\sigma(\beta_e(1-d_e))$ with
$\beta_e>0$, this is equivalent to $d_e\le1$. Thus edge testing requires neither
the value of $\beta_e$ nor a fitted Mahalanobis matrix. Under other response
models, we take the majority-response rule itself to define the target
specification.

\paragraph{Confidence bands under adaptive allocation and stopping.}
Inspect the response stream of each queried pair at sample sizes
$N_j=2^j$, $j=1,2,\ldots$. Let $\widehat p_{e,j}$ be its sample mean and set
\begin{equation}\label{eq:direct_width_v6}
 b_j=\sqrt{\frac{\log(2m j(j+1)/\delta)}{2N_j}},\qquad
 I_{e,j}=[\widehat p_{e,j}-b_j,\widehat p_{e,j}+b_j]\cap[0,1].
\end{equation}
Intersect successive intervals for a pair, using $[0,1]$ before its first
inspection; if the intersection is empty, abstain. Write $[L_e,U_e]$ for the
current interval. Hoeffding's inequality and
$\sum_{j\ge1}1/(j(j+1))=1$ give
\begin{equation}\label{eq:direct_event_v6}
 \Pr\{p_e\in I_{e,j}\text{ for all }e,j\}\ge1-\delta.
\end{equation}
The union bound covers prefixes of the response streams, so choosing
which pair to query next, and when to stop at these inspection times,
preserves coverage. The bound also permits dependence across pairs,
provided each pair's prefix concentration assumption remains valid.

Build the possible-edge (safe) graph from $U_e\ge1/2$ and the
certain-edge (relaxed) graph from $L_e\ge1/2$, both with budget
$\epsilon_e$. On the event above, their
feasible classes bracket the target class at every inspection time.
The repair-radius certificate of Theorem~\ref{thm:uncertain_repair}
can therefore certify feasibility or impossibility at any inspection
time. Every feasibility certificate includes a repaired score vector.
If the budget remains unresolved, the procedure continues querying or
abstains. Pair selection may use the current graph and movement budget.

\paragraph{Response margin for budget certification.}
For $0<\gamma\le1/2$, define two deterministic graphs using the unknown
response means:
\[
 E_\gamma^{\rm certain}=\{e:p_e\ge1/2+\gamma\},\qquad
 E_\gamma^{\rm possible}=\{e:p_e\ge1/2-\gamma\},
\]
with the same known tolerances. These graphs define a sufficient condition
for budget resolution; the procedure operates on the confidence
bands above.

\begin{theorem}[Budget certification without complete pair identification]
\label{thm:direct_budget_v6}
Suppose at some $\gamma>0$ either
\begin{equation}\label{eq:decision_margin_v6}
 r_{G_\gamma^{\rm possible}}(h)\le t
 \quad\text{or}\quad
 r_{G_\gamma^{\rm certain}}(h)>t.
\end{equation}
On the event~\eqref{eq:direct_event_v6}, the procedure resolves the budget
once every still-unclassified pair has been sampled through a common
dyadic inspection $j$ satisfying $2b_j<\gamma$. Classified pairs need no further
responses. Thus round-robin sampling of unclassified pairs uses at most
$m2^{j_\gamma}$ responses, where
\[
 j_\gamma=\min\{j\ge1:
 2^j>2\gamma^{-2}\log(2m j(j+1)/\delta)\}.
\]
The response cost is
$O\big(m\gamma^{-2}[\log(m/\delta)+\log\log(e/\gamma)]\big)$.
Condition~\eqref{eq:decision_margin_v6} can hold even when some
$p_e=1/2$, so the procedure can decide the budget question without classifying
every pair.
\end{theorem}

\begin{proof}
Coverage at the current inspection implies that the lower endpoint is at least
$p_e-2b_j$ and the upper endpoint is at most $p_e+2b_j$. Every pair with
$p_e\ge1/2+\gamma$ is therefore certified present, while every pair with
$p_e<1/2-\gamma$ is certified absent. Previously classified pairs are
correct on the same event. Hence the current relaxed graph contains
$G_\gamma^{\rm certain}$ and the safe graph is a subgraph of
$G_\gamma^{\rm possible}$. Adding constraints increases the repair radius,
so either inequality in~\eqref{eq:decision_margin_v6} forces the
corresponding stopping condition. The first qualifying dyadic inspection has
the stated order: its confidence penalty grows as $\log(m/\delta)+2\log j$,
whereas $N_j$ grows geometrically. Every sampled pair contributes at most
$2^{j_\gamma}$ responses, including all previous prefixes.
\end{proof}

At one prespecified sample size, the simpler width
$b=\sqrt{\log(2m/\delta)/(2N)}$ gives the same conclusion whenever
$N>2\gamma^{-2}\log(2m/\delta)$. Adaptive stopping uses the dyadic
confidence bands instead. Here $m$ counts pairs and $N$ counts responses
\emph{per pair}, so the total response cost scales with the candidate-pair
set. Appendix~\ref{app:coupled_queries_v9} uses known response-equivalence
classes to share information across pairs in the sharp metric family.

\paragraph{Certification from a subset of pairs.}
Let a nonempty $S\subseteq\mathcal P$ be fixed before observing responses,
and suppose its true edge labels alone determine the budget answer when
every pair outside $S$ is left possible but not certain. If
$\min_{e\in S}|p_e-1/2|>0$, the preceding guarantees apply with $m=|S|$
and any smaller positive margin, querying only $S$; all unqueried
intervals remain $[0,1]$. If the empty set already determines the answer,
no responses are needed. For an infeasible budget, a sufficient certificate
is a path of certified-present edges whose total budget is smaller than
$|h_u-h_v|-2t$. For a feasible budget, one may supply $z$ within budget
and certify absence of every candidate pair on which
$|z_u-z_v|>\epsilon_e$. Both certificates can be checked directly. Finding a certificate that
uses few pairs may require additional queries; adaptively choosing $S$ requires confidence allocation covering
all pairs that the search can inspect.

\paragraph{Example: a single-pair certificate.}
Consider a matching of $m$ candidate pairs with tolerance $0.1$.
Scores on pair 1 are $(0,1)$, while every other score is $1/2$.
Let $p_1=0.75$ and $p_e=0.5$ on the other pairs. Only pair 1 can
require movement: the initial radius interval is $[0,0.45]$, and
certifying it present fixes the repair radius at $0.45$, simultaneously
resolving every movement budget. The repaired pair is $(0.45,0.55)$;
all other scores remain unchanged. Because only this prespecified pair is queried, the confidence bound
covers one pair, with response margin $1/4$. Although all remaining
response means equal one half, their edge labels do not affect the
budget decision.

\begin{proposition}[Response-margin lower bound for budget certification]
\label{prop:direct_lower_v6}
On two profiles with $h=(0,1)$ and known $0\le\epsilon<1$, fix
$0\le t<(1-\epsilon)/2$. Compare similarity means
$p_+=1/2+\gamma$ and $p_-=1/2-\gamma$, $0<\gamma<1/2$.
Any sequential procedure that terminates with the correct budget answer
with probability at least $1-\delta$ under both specifications,
$0<\delta<1/2$, has
\[
 \mathbb E_+ N\ \ge
 \frac{\operatorname{kl}(1-\delta,\delta)}
 {\operatorname{kl}(1/2+\gamma,1/2-\gamma)}
 =\frac{(1-2\delta)\log((1-\delta)/\delta)}
 {2\gamma\log((1/2+\gamma)/(1/2-\gamma))}.
\]
The same bound holds for $\mathbb E_-N$. In particular, this is
$\Omega(\gamma^{-2}\log(1/\delta))$ for $\gamma\le1/4$ and
$\delta\le1/10$.
\end{proposition}
\begin{proof}
The specification with mean $p_+$ requires radius $(1-\epsilon)/2$,
whereas the specification with mean $p_-$ requires radius zero, so the
correct answers differ.
If the expected sample size is infinite, the claim is immediate. Otherwise,
the stopped-transcript likelihood-ratio identity gives
$\operatorname{KL}(P_+^{\rm transcript},P_-^{\rm transcript})
=\mathbb E_+N\operatorname{kl}(p_+,p_-)$.
Data processing to the event ``impossible'' lower-bounds this divergence
by $\operatorname{kl}(1-\delta,\delta)$. Interchanging the two laws gives
the reverse bound. Expanding Bernoulli KL yields the displayed expression.
This two-point testing argument gives the response-margin dependence
for a single pair. The construction also lies in the logistic
distance model when its known $\beta$ is chosen so both distances
$1-\operatorname{logit}(p_\pm)/\beta$ are nonnegative.
\end{proof}

\subsection{Two response streams for the alternating construction}
\label{app:coupled_queries_v9}

Known metric coupling can reduce the number of response streams.
Pairs with the same outer product
$(x_u-x_v)(x_u-x_v)^\top$ have equal Mahalanobis distance for every
matrix. Under a common logistic response slope, they also have equal
response probabilities, so one representative provides information about the entire class.

\begin{proposition}[Query complexity with known response-equivalence classes]
\label{prop:class_queries_v9}
Suppose the pairs not already classified by known geometry fall into
$q\ge1$ known classes. Class $j$ has common response mean $p_j$ and
is constrained exactly when $p_j\ge1/2$. Tolerances are fixed and known.
Query one prespecified representative per class, with independent
Bernoulli responses within its stream. At a fixed sample size $N$ per
representative, form
\[
 I_j=[\widehat p_j-b_N,\widehat p_j+b_N]\cap[0,1],
 \qquad b_N=\sqrt{\frac{\log(2q/\delta)}{2N}}.
\]
Classify the class as present if $\inf I_j\ge1/2$, absent if
$\sup I_j<1/2$, and unresolved otherwise. With probability at least
$1-\delta$, every classified label is correct. If all classes are
classified, the graph-envelope repair is optimal under the true
specification; otherwise the procedure returns a certificate from the
safe and relaxed graphs or abstains.

If $\min_j|p_j-1/2|\ge a>0$, all classes are classified on this
coverage event whenever
\begin{equation}\label{eq:class_query_count_v9}
 N>2a^{-2}\log(2q/\delta).
\end{equation}
Under a common logistic slope $\beta>0$ and
$|d_j-1|\ge\kappa>0$, one may take
\begin{equation}\label{eq:logistic_class_margin_v9}
 a=\tfrac12\tanh(\beta\kappa/2),
\end{equation}
yielding total response cost
$O(q\max\{1,(\beta\kappa)^{-2}\}\log(q/\delta))$.
\end{proposition}
\begin{proof}
Hoeffding's inequality and a union bound give simultaneous coverage of
the $q$ means. Each classified label is then correct for its entire class,
and the known graph and budgets determine the exact repair.
On coverage, $\inf I_j\ge p_j-2b_N$ and
$\sup I_j\le p_j+2b_N$, with clipping only improving these bounds.
Thus $2b_N<a$ ensures every label is classified, giving the sample
condition. Logistic symmetry gives the margin formula, and
$\tanh t$ is of order $\min\{t,1\}$ for $t>0$.
\end{proof}

For adaptive allocation and stopping, use the dyadic bands in
\eqref{eq:direct_width_v6} with confidence-family size $q$.
The resulting bound has the additional $\log\log(e/a)$ term; threshold
ties may remain unresolved, while every returned certificate remains valid
on the simultaneous coverage event.

\begin{corollary}[Optimal repair with a response count independent of pool size]
\label{cor:elicitable_sharpness_v9}
Use the alternating profiles and ball
$\mathcal C=\{M:\|M-2I_2\|_F\le7/5\}$ of
Proposition~\ref{prop:alternating_geometry_v9}, with $n\ge3$.
The representation, ball, and common tolerance are known;
$M^\star\in\mathcal C$ is unknown. Responses obey
$p_e=\sigma(\beta[1-d_{M^\star}(e)])$ with known common slope
$\beta>0$ and fresh independent randomness within each pair stream.
Querying only $(1,2)$ and $(2,3)$ returns, with probability at least
$1-\delta$, an optimal repair or an abstention.
At either interior matrix $\operatorname{diag}(2/3,2)$ or
$\operatorname{diag}(2,2/3)$, at most $2N$ responses suffice for an
optimal repair with probability at least $1-\delta$, where
\begin{equation}\label{eq:two_stream_query_count_v9}
 \kappa_0=1-\sqrt{2/3},\quad
 a_0=\tfrac12\tanh(\beta\kappa_0/2),\quad
 N=\left\lfloor2a_0^{-2}\log(4/\delta)\right\rfloor+1.
\end{equation}
The count is independent of $n$. For ramp scores and tolerance
$\eps=\gamma/(n-1)$, $0\le\gamma<1$, the returned movement is
$(1-\gamma)/[2(n-1)]$, compared with common repair cost $(1-\gamma)/2$.
Graph construction and repair take $O(n)$ time and storage.
\end{corollary}
\begin{proof}
Odd and even adjacent pairs have squared distances $M_{11}$ and
$M_{22}$, respectively. The spectral lower bound and profile geometry
exclude every nonadjacent pair, leaving two response-equivalence classes.
At these matrices, their distances are $\sqrt{2/3}$ and $\sqrt2$,
both at least $\kappa_0$ from one. The preceding proposition with
$q=2$ proves the response guarantee.

The costs follow from Proposition~\ref{prop:alternating_geometry_v9}.
Explicitly, put $r=(1-\gamma)/[2(n-1)]$ and move every active pair
$(i,i+1)$ to $(h_i+r,h_{i+1}-r)$, leaving other scores unchanged.
The pairs are disjoint, the new scores remain in $[0,1]$, and each gap
is exactly the known tolerance. Generating the matching and these
assignments takes $O(n)$ time and storage.
\end{proof}

The off-diagonal matrix entry need not be identified. The geometry
also permits an earlier decision: once either orientation is certified
present, the other is necessarily absent, since both diagonals cannot be
at most one
in $\mathcal C$. The pool-independent response bound holds at the
interior matrices; elsewhere in the ball it depends on the activation
margin through~\eqref{eq:logistic_class_margin_v9}.

\section{Critical paths and tolerance information on general graphs}
\label{app:tolerance_geometry}

Let $S=(V,E_S)$ have fixed finite nonnegative budgets $s_e$, and let
$T=(V,E_T)$ have the common budget $\eps\ge0$ on each edge. The
multigraph $\mathcal G$ includes parallel source and target copies of
every shared edge. Satisfying its parallel edges is equivalent to using the smaller
budget on the overlap. Throughout, the adaptation interpretation assumes
$h\in\cF_S$; the path identities below also hold for arbitrary
$h\in[0,1]^n$. Paths are vertex-simple and retain their edge-copy choices.
There are finitely many such paths, even when some budgets are zero.

\subsection{Path envelope and exact tolerance sensitivity}

\begin{proof}[Proof of Theorem~\ref{thm:tolerance_path_geometry}]
Write $d_\eps$ for shortest-path distance in $\mathcal G$, with source
lengths $s_e$ and target lengths $\eps$. All lengths are nonnegative,
so each connected pair has a simple shortest path, and
\[
 d_\eps(u,v)=\min_{P:u\leadsto v}(A_P+k_P\eps).
\]
For connected pairs, the cap at one in the exact repair identity does
not affect the positive deficit because $|h_u-h_v|\le1$.
Disconnected pairs contribute zero. Substituting the shortest-path
minimum into the radius identity and interchanging finite maxima proves
\eqref{eq:tolerance_path_envelope}.

Each line has nonpositive slope $-k_P/2$, and the additional constant
zero line enforces nonnegativity. The maximum is therefore
nonincreasing, convex, continuous, and piecewise affine. Suppose first
that $r(a)>0$. The zero line is then inactive. Among the finitely many
lines attaining $r(a)$, the largest slope is
$-\min k_P/2=-\kappa_{\rm tol}(a)/2$. Every inactive line has a
strictly positive gap below $r(a)$. Finiteness ensures that, on some
interval immediately to the right of $a$, no inactive line overtakes
the active line with largest slope. Hence
\[
 r(a+t)=r(a)-\frac{\kappa_{\rm tol}(a)}2t
 \quad\text{for all sufficiently small }t\ge0.
\]
This proves the right derivative and its local exactness, including a
tie between critical paths with different target-edge counts.

The slopes of a convex piecewise-affine function are nondecreasing.
Every slope to the right of $a$ consequently lies between
$-\kappa_{\rm tol}(a)/2$ and zero. Summing the changes over its affine
pieces proves~\eqref{eq:tolerance_critical_sensitivity} for
$a\le x\le y$. If $r(a)=0$, nonnegativity and monotonicity give
$r(\eps)=0$ for all $\eps\ge a$, proving the same conclusions with
$\kappa_{\rm tol}(a)=0$.

A simple path has at most $n-1$ edges. If $h\in\cF_S$, a path containing
only source edges satisfies $H_P\le A_P$ by the triangle inequality
along the path. Such a path cannot attain a positive repair deficit.
Thus $1\le\kappa_{\rm tol}(a)\le n-1$ whenever $r(a)>0$.
The algorithm below proves the computational claim.
\end{proof}

If source feasibility is dropped, a source-only path can be critical
with $\kappa_{\rm tol}(a)=0$ and $r(a)>0$. The conclusions remain valid:
that path supplies a constant lower bound equal to $r(a)$, while
monotonicity supplies the same upper bound for every $\eps\ge a$.
The radius is then constant, and a repair at $a$ is optimal throughout
the remaining tolerance range. Under the source-feasible assumption,
this case is excluded by the preceding proof.

\paragraph{Computing the critical target-edge count.}
At tolerance $a$, initialize every vertex $v$ as a Dijkstra source with
the lexicographic label $(h_v,0)$. A source edge adds $(s_e,0)$ and a
target edge adds $(a,1)$. Compare labels by their first coordinate,
breaking exact ties by the second. Nonnegative increments permit the
usual Dijkstra search with these ordered labels. Its final label at
$u$ is $(D_u,C_u)$, where
\[
 D_u=\min_{v,P:v\leadsto u}\{h_v+A_P+k_Pa\},
 \qquad
 C_u=\min\{k_P:h_v+A_P+k_Pa=D_u\}.
\]
The empty path is allowed and has count zero. Any minimizing walk can
be made simple: removing a cycle decreases the first coordinate, or
leaves it unchanged and does not increase the second. Zero-length
source-only cycles can be removed without changing either coordinate.
This argument also covers zero budgets.

Undirectedness lets us orient any endpoint pair from its lower source
score to its higher source score. The radius is therefore
\[
 2r(a)=\max_u(h_u-D_u).
\]
The maximum is nonnegative because $D_u\le h_u$. If it is positive,
\[
 \kappa_{\rm tol}(a)=
 \min\{C_u:h_u-D_u=2r(a)\}.
\]
Indeed, a pair and path attaining the radius must minimize
$h_v+\operatorname{length}(P)$ at its higher-score endpoint; otherwise
a shorter such label would produce a larger deficit. Conversely, a
label attaining the displayed maximum gives a critical path.
Predecessors recover a path with the indicated count; ties with
unchanged labels need not replace predecessors. The search uses
$O((|E_S|+|E_T|+n)\log n)$ time and $O(|E_S|+|E_T|+n)$ storage for
$n\ge2$, in the same arithmetic model as ordinary weighted Dijkstra.
No enumeration of all simple paths is required. The case $n=1$ has
radius zero and requires no queries.

\subsection{Movement certificates from heterogeneous tolerance bands}

For each target edge, let its true budget be $c_e^\star$ and suppose
$0\le\ell_e\le c_e^\star\le u_e$. Source-edge budgets remain fixed.
Let $r_\ell$ be the radius on the graph of lower budgets, and let
$r_\star$ be the true radius. The lower-budget optimum is feasible for
the true graph, so $r_\star\le r_\ell$. If $r_\ell>0$, choose a
critical path $P$ of the lower-budget graph. The true shortest-path
distance between its endpoints is at most the true length of this
particular path. Therefore
\begin{align*}
 2r_\star
 &\ge H_P-A_P-\sum_{e\in P\cap E_T}c_e^\star\\
 &\ge H_P-A_P-\sum_{e\in P\cap E_T}\ell_e
       -\sum_{e\in P\cap E_T}(u_e-\ell_e)\\
 &=2r_\ell-\sum_{e\in P\cap E_T}(u_e-\ell_e).
\end{align*}
The first inequality remains true when its right side is negative.
This proves~\eqref{eq:tolerance_path_width}. If $r_\ell=0$, the
original scores satisfy the lower-budget graph and hence the true one.

Put $W_P=\sum_{e\in P\cap E_T}(u_e-\ell_e)$. The inequality
$W_P\le2\eta$ gives excess movement at most $\eta$.
If $W_P\le2\alpha r_\ell/(1+\alpha)$, it instead gives
$r_\star\ge r_\ell/(1+\alpha)$, proving the multiplicative guarantee.
When a community tolerance is shared by several edges, each edge on
the path contributes its occurrence of the same interval width. No
independence between the edge bands is required.

These are deterministic implications of simultaneous band coverage.
If coverage also holds across every inspected sample size, they remain
valid when the queried streams, critical paths, and stopping time are
selected from the responses. One valid rule repeatedly computes a safe
critical path and queries unresolved tolerance streams on that path
until a certificate holds, another path becomes critical, or the
procedure abstains. A query-complexity bound for this allocation rule would also require
controlling how many critical paths it visits. Its confidence bands must
cover all streams that the adaptive search can inspect.

\subsection{Fixed-sample upper bound}

\begin{proof}[Upper bound in Theorem~\ref{thm:tolerance_general_queries}]
The prior interval has $0\le a<b\le1$, so its midpoint
$g_*=(a+b)/2$ is a permitted displayed gap. Under the response model,
fresh responses at this gap are Bernoulli with mean $p_\eps(g_*)$.
Let $N$ be~\eqref{eq:tolerance_general_upper}, let $\widehat p$ be the
sample mean, and set
\[
 B_N=\sqrt{\frac{\log(2/\delta)}{2N}}.
\]
Intersect $[\widehat p-B_N,\widehat p+B_N]$ with
$[p_a(g_*),p_b(g_*)]$. Abstain if the intersection is empty.
Otherwise denote its endpoints by $L,U$ and invert them to
\[
 \ell=g_*+\frac{\operatorname{logit}(L)-\operatorname{logit}(q)}\lambda,
 \qquad
 u=g_*+\frac{\operatorname{logit}(U)-\operatorname{logit}(q)}\lambda.
\]
Both endpoints lie in $[a,b]$. The response probabilities at the prior
interval's endpoints lie strictly between zero and one, so these
inversions are finite.
Throughout this probability interval the inverse derivative is at most
$1/(\lambda v)$. On the Hoeffding coverage event, of probability at
least $1-\delta$, the intersection is nonempty,
$\ell\le\eps\le u$, and
\[
 u-\ell\le\frac{2B_N}{\lambda v}.
\]
Return the graph-envelope repair at tolerance $\ell$, retaining every
source edge. It is feasible under the true tolerance because
$\ell\le\eps$. Its movement is exactly $r(\ell)$, and
Theorem~\ref{thm:tolerance_path_geometry} gives
\[
 r(\ell)-r(\eps)
 \le\frac{\kappa_{\rm tol}(a)}2(\eps-\ell)
 \le\frac{\kappa_{\rm tol}(a)B_N}{\lambda v}
 \le\alpha r_{\min}\le\alpha r(\eps).
\]
Here $r_{\min}=r(b)>0$ and source feasibility ensure
$\kappa_{\rm tol}(a)\ge1$, so the stated positive sample count and
all divisions are valid.
\end{proof}

Replacing $\alpha r_{\min}$ by a desired additive error $\eta>0$
gives the sample bound
\[
 N=\left\lceil
 \frac{\kappa_{\rm tol}(a)^2\log(2/\delta)}
 {2\lambda^2v^2\eta^2}
 \right\rceil,
\]
without assuming $r(b)>0$. If $r(a)=0$, no responses are needed and
the procedure returns $h$. If $a=b$, the tolerance is known and its
optimal repair likewise needs no tolerance responses.

More generally, any valid common-tolerance interval $[\ell,u]$ permits
the exact multiplicative stopping test
$r(\ell)\le(1+\alpha)r(u)$, without replacing the curve by a uniform
sensitivity bound. Repairing at $\ell$ is then feasible and has
movement at most $(1+\alpha)r(\eps)$. In particular, if this holds for
the entire prior interval, no tolerance responses are needed. When
$r(a)>(1+\alpha)r(b)$ and $\delta<1/2$, no response-free randomized
procedure can satisfy the guarantee at both endpoints: feasibility
at $a$ and near-optimality at $b$ require disjoint ranges of movement,
whereas a response-free procedure has the same output law at both.

If the known target graph was itself obtained from similarity queries,
allocate separate failure probabilities to graph identification and to
fresh tolerance responses. Conditional on the identified graph and
past observations, the latter obey the stated response law. The union
bound yields the joint guarantee on correct graph identification and
tolerance coverage. This theorem neither identifies unresolved
edges nor bounds the cost of that separate graph-learning task.

\subsection{Local lower bound}

\begin{proof}[Lower bound in Theorem~\ref{thm:tolerance_general_queries}]
Both tolerances lie in the stated affine interval, so
\[
 r(\eps_1)=r_0-\frac k2\frac{2\alpha r_0}k
 =(1-\alpha)r_0.
\]
Let $A$ be the event of returning a repair whose movement is at least
$r_0$; abstentions do not belong to $A$. At $\eps_0$, feasibility for
the source and target together forces movement at least $r_0$.
Thus $\Pr_0(A)\ge1-\delta$. At $\eps_1$, every successful return has
movement at most
\[
 (1+\alpha)r(\eps_1)=(1-\alpha^2)r_0<r_0,
\]
so $\Pr_1(A)\le\delta$.

At any displayed gap, the two Bernoulli natural parameters differ by
$\lambda(\eps_1-\eps_0)=2\lambda\alpha r_0/k$. For Bernoulli
variables, the relative entropy is a Bregman remainder of
$F(t)=\log(1+e^t)$. Since $F''(t)\le1/4$, either direction of the
one-response divergence is at most
\[
 \frac{\lambda^2(\eps_1-\eps_0)^2}{8}
 =\frac{\lambda^2\alpha^2r_0^2}{2k^2}.
\]
This holds uniformly over all permitted adaptive gap choices.
Action-selection kernels coincide conditional on each history, and
other observations with identical conditional laws contribute no
divergence. The standard stopped-transcript change-of-measure
argument~\citep{kaufmann2016complexity}, followed by data processing
to $A$ in direction zero and $A^c$ in direction one, gives
\[
 \mathbb E_iN_{\rm tol}
 \frac{\lambda^2\alpha^2r_0^2}{2k^2}
 \ge\operatorname{kl}(1-\delta,\delta),\qquad i\in\{0,1\}.
\]
For finite expectations one may first apply the argument to truncated
transcripts and then pass to the almost surely terminated transcript;
infinite expectations satisfy the bound directly. Rearrangement proves
\eqref{eq:tolerance_general_lower}.
\end{proof}

The local assumption is substantive: the interval must extend at least
$2\alpha r_0/k$ to the right of $\eps_0$ without changing the affine
radius formula. Finiteness of the path envelope supplies some positive
right interval whenever $r_0>0$ and its right slope is nonzero, but that
interval may be short. On a prior interval where the radius is positive and affine, $\kappa_{\rm tol}(a)=k$. If its radius stays within constant
factors of $r_0$ and it contains the two lower-bound alternatives, the
upper and lower sample bounds have the same dependence on $k^2\alpha^{-2}r_0^{-2}\log(1/\delta)$
for fixed $q,\lambda$ and $\delta\le1/10$.
For an identified matching in the alternating construction,
$k=1$ and $r_0$ is of order $1/(n-1)$, recovering the quadratic pool-size
rate. Other resolved graphs, including those retaining source edges,
are governed by their own critical target-edge counts and repair radii.

\section{Tolerance precision in the alternating-matching family}
\label{app:tolerance_tradeoff_v11}

We specialize the general tolerance results to the alternating family of
Proposition~\ref{prop:alternating_geometry_v9}. Let $n\ge3$,
$\Delta=1/(n-1)$, and $h_i=(i-1)\Delta$. The two nonempty target
graphs are matchings whose union is the full path; every adjacent score
gap is $\Delta$. All active edges share tolerance $\eps$.

\begin{proposition}[Matching and common repair costs under tolerance uncertainty]
\label{prop:tolerance_interval_v11}
For an identified nonempty matching and for the union path, respectively,
\[
 r_c=\tfrac12[\Delta-c]_+,
 \qquad R_c=\tfrac12[1-(n-1)c]_+=(n-1)r_c.
\]
A supplied interval $0\le\ell\le\eps\le u$ gives exact cost intervals
$[r_u,r_\ell]$ and $[R_u,R_\ell]$. Repairing at $\ell$ minimizes
movement subject to feasibility for every tolerance in the interval,
with excess cost
\[
 0\le r_\ell-r_\eps\le\tfrac12\min\{u-\ell,\Delta\}.
\]
Equalizing the matching's active pairs uses movement $\Delta/2$ and
is feasible for every $\eps\ge0$. If $\eps\le\Delta/2$, it is a
$2$-approximation, and common repair at the true tolerance costs at least
$(n-1)/2$ times as much movement.
\end{proposition}
\begin{proof}
On a matching edge, movement $r$ reduces a gap by at most $2r$;
symmetric inward movement attains $[\Delta-c]_+/2$ independently on
all active pairs. On the common path, vertices $k$ edges apart have
score gap $k\Delta$ and path budget $kc$, so the endpoint pair gives
$R_c=(n-1)r_c$. Monotonicity and the Lipschitz property of the
positive part give the interval and excess bounds. Equalization is
repair at $c=0$; for $\eps\le\Delta/2$, $r_\eps\ge\Delta/4$ and
$R_\eps\ge(n-1)\Delta/4$, proving both comparisons.
\end{proof}

\begin{proof}[Proof of Theorem~\ref{thm:tolerance_tradeoff_main_v11}]
Corollary~\ref{cor:elicitable_sharpness_v9} identifies the matching
from two response streams at either interior target matrix, with a
response count independent of $n$. The preceding proposition then gives
the feasible $2$-approximation without tolerance responses.

For nearly optimal movement, use the fixed-slope response model
\eqref{eq:tolerance_oracle_geometry}. On the prior interval
$[0,\Delta/2]$, the matching has $\kappa_{\rm tol}(0)=1$ and
$r_{\min}=\Delta/4$. The upper bound of
Theorem~\ref{thm:tolerance_general_queries}, at gap $\Delta/4$, thus
uses at most
\[
 N=\left\lceil
 \frac{8\log(4/\delta)}{\lambda^2v^2\alpha^2\Delta^2}
 \right\rceil,
 \qquad
 v=\min_{|s|\le\lambda/8}
 \sigma(\operatorname{logit}(q)+s)
 [1-\sigma(\operatorname{logit}(q)+s)]>0.
\]
The constant is uniform in $n$ since $\Delta\le1/2$.
Allocate $\delta/2$ to graph identification and $\delta/2$ to fresh
tolerance responses. The union bound gives the joint guarantee;
on unresolved graphs the procedure abstains.

For the explicit lower-bound constant, take
$\eps_0=\Delta/4$ and $\eps_1=(1/4+\alpha)\Delta$, both in
$[0,\Delta/2]$ for $0<\alpha\le1/4$.
Feasibility at $\eps_0$ requires movement at least $3\Delta/8$,
whereas success at $\eps_1$ requires movement at most
\[
 (1+\alpha)r_{\eps_1}
 =\frac{3\Delta}{8}-\frac\Delta2(\alpha/4+\alpha^2)
 <\frac{3\Delta}{8}.
\]
The event of returning movement at least $3\Delta/8$ therefore has
probabilities at least $1-\delta$ and at most $\delta$, respectively;
abstentions are outside this event. At any queried gap, the response
natural parameters differ by $\lambda\alpha\Delta$. Since the
Bernoulli log-partition function has second derivative at most $1/4$,
the one-response divergence in either direction is at most
$\lambda^2\alpha^2\Delta^2/8$. The stopped-transcript argument in
Appendix~\ref{app:tolerance_geometry} consequently gives
\[
 \mathbb E_iN_{\rm tol}\ge
 \frac{8\operatorname{kl}(1-\delta,\delta)}
 {\lambda^2\alpha^2\Delta^2},\qquad i\in\{0,1\}.
\]
The bound permits adaptive gaps, internal randomization, and interleaved
responses whose conditional laws coincide under the two tolerances.
Infinite expected counts satisfy the inequality directly. Substituting
$\Delta=1/(n-1)$ proves the displayed lower bound and matching rate.
\end{proof}

\section{Elicitation and training protocols}
\label{app:experiments}

This appendix specifies the COMPAS studies and numerical comparisons in
Section~\ref{sec:experiments}. The studies examine how respondents specify
similarity and acceptable score gaps. The training comparison examines the
relationship between compliance with fixed numerical constraints and
predictive utility.

\subsection{Similarity and tolerance judgments}

\paragraph{Profiles and similarity judgments.}
We use ProPublica's two-year recidivism data~\citep{compas}, restricted to
African-American and Caucasian defendant subsets. In the similarity study,
10 annotators judge 500 pairs per community. Profile cards display age,
sex, charge degree and description, and criminal history; race is omitted
from the cards. The task heading identifies the defendant population.
Thus a profile's community identifies the population being judged, while
the composition of the respondent panel is a separate design choice.
The instructions state:
\begin{quote}
For each pair of defendant profiles, decide whether a recidivism risk
assessment system should treat them \emph{similarly}. ``Similarly situated''
means the two defendants are comparable enough that assigning
significantly different risk scores would be unfair.
\end{quote}
The question is ``Should a risk assessment system treat these two defendants
similarly?'' Respondents choose ``Similarly Situated'' or ``Not Similarly
Situated''. The fitting coordinates are age, prior count, three
juvenile-history counts, charge degree, and sex, standardized over the
filtered dataset. For each community, we fit the working model
$p_M(x,x')=\sigma(5[1-d_M(x,x')])$, where
$\sigma(t)=(1+e^{-t})^{-1}$, to majority-vote similarity labels, with
ties assigned to ``Not Similarly Situated''.
We parameterize $M=L^\top L+10^{-3}I$ and minimize the summed Bernoulli
negative log-likelihood plus $10^{-3}\|L\|_F^2$, using five L-BFGS-B
restarts. Fleiss' $\kappa$ is $.549$ and $.512$ in the two communities.

\paragraph{Tolerance study.}
Thirty Black/African-American respondents and thirty White/Caucasian
respondents, recruited using Prolific demographic prescreens, each judge
200 pairs from the corresponding defendant community, giving 12,000
responses. Pairs lie within the fitted community unit balls and are
stratified into five score-gap bins. A logistic-regression scorer supplies
the displayed risks. Cards show the two profiles, their risk scores, and
the absolute score difference; race is omitted. The instructions explain:
\begin{quote}
Each pair below shows two defendants the community's similarity standard
considers \emph{similarly situated} (Phase 1 already confirmed this).
A risk-assessment model has scored each defendant. Decide whether the
\emph{difference} between the two scores is acceptable for
similarly-situated individuals.

\textbf{Acceptable} = the score difference is small enough that we are
comfortable deploying this model for these two defendants.
\textbf{Not acceptable} = the score difference is large enough that the
model is treating similar people unfairly.
\end{quote}
The question is ``Is this score difference acceptable for similarly-situated
defendants?'' The response options are ``Acceptable'' and ``Not Acceptable''.

\paragraph{Response summaries.}
Table~\ref{tab:phase2} reports two distinct summaries of these judgments.
For each respondent, a logistic fit models acceptance as a function of
the displayed gap, with inverse regularization parameter $C=10^6$ and
either sign permitted for the slope. The fitted 50\% acceptance point is
$-\mathrm{intercept}/\mathrm{slope}$, clipped to $[0,1]$.
The cumulative-grid heuristic scans cutoffs $.01,.02,\ldots,1$ in order.
At each cutoff it requires at least five observations and at least 90\%
acceptance among observations below the cutoff, and stops at the first
failure. If no cutoff passes, it returns the minimum observed gap.
The table gives medians of the respondent-specific estimates, together
with the grid interquartile ranges. Its acceptance column is the overall
fraction of judgments marked acceptable. The logistic and grid summaries
use different acceptance criteria.

\paragraph{Study protocol.}
The studies followed a single IRB-approved protocol. Participants gave
informed consent and received compensation at \$20/hour. The consent
screen described the study purpose, time commitment, anonymity,
withdrawal rights, and payment. The tasks consisted of minimal-risk survey
judgments. The quoted instructions and aggregate summaries specify the
study procedures and reported outcomes; participant-level responses and
original recruitment and approval records are not included in the
accompanying materials.

\subsection{Training and score compression}

\paragraph{Data and feature representation.}
The training comparison uses the filtered COMPAS data and the same two
defendant subsets. Its seven features are age, prior count, three
juvenile-history counts, and the recidivism and violent-recidivism decile
scores. Standardization is fitted on the full filtered dataset before
the outer split. The last two coordinates therefore differ from those
used to elicit the matrices and tolerances. The matrices define fixed
numerical quadratic forms in this training representation; their
elicited interpretation applies to the fitting representation above.
We write $d_k$ for the resulting numerical distance in community $k$.

\paragraph{Training configurations.}
Table~\ref{tab:common_protocol_v5} compares two penalty forms at three
training-tolerance settings: U uses $(.10,.10)$, H uses $(.039,.063)$,
and M uses $(.039,.039)$. The H values are the reported logistic
medians, treated as fixed numerical specifications in this comparison.
Every configuration uses a sigmoid-output MLP with two 64-unit ReLU
hidden layers and dropout $.2$, predictive binary cross-entropy, and a
pairwise penalty with weight $\lambda=2$.
The Lipschitz penalty averages
$[|h(x)-h(x')|-\eps_kd_k(x,x')]_+$ over sampled pairs with $d_k\le3$.
The threshold penalty averages
$[|h(x)-h(x')|-\eps_k]_+$ over sampled pairs with $d_k\le1$.

Outer train/test splits use 70/30 proportions and random states
$s=0,1,2,3,4$; the same $s$ seeds model training. We use Adam with learning
rate $10^{-3}$, batches of 256, and at most 40 epochs. A seeded subset of
$\min(\lfloor .15n_{\rm outer\text{-}train}\rfloor,2000)$ records is
reserved for predictive validation loss. Training stops after eight
epochs without improvement exceeding $10^{-4}$ and restores the best
checkpoint. Computations use standard CPU hardware with eight threads.

The Lipschitz implementation samples up to 64 pairs per community within
each supervised minibatch. The threshold implementation draws 64 pairs
per community from the entire outer-training community pool for each
minibatch, then removes self-pairs and applies the distance cutoff. Its
penalty pool includes the feature rows reserved for predictive early
stopping; their outcome labels are used only for validation. It makes
separate stochastic model calls for the two penalty endpoints. The
comparison consequently reflects penalty form, pair sampling, and
stochastic evaluation together. Both procedures optimize soft penalties.

\paragraph{Score compression.}
For the Lipschitz-U base, we also evaluate
$h_\alpha=\tfrac12+\alpha(h-\tfrac12)$ at
$\alpha\in\{1,.75,.5,.25,.1\}$. This transformation preserves decisions
at threshold $.5$ and multiplies every score gap by $\alpha$. The base
is retrained with the same seed before evaluating all five factors.

\paragraph{Common evaluation.}
All configurations and compression factors are evaluated at tolerances
$(.039,.063)$. For each seed, the evaluator draws 20,000 pairs per
community, removes self-pairs, and conditions on $d_k\le1$. Compression
evaluation resets the random generator to the same seed and uses the
same community order and sampling procedure, giving identical evaluation
pairs within each seed. Strict violation is the equally weighted average
of the two within-community fractions whose score gaps exceed their
evaluation tolerances. Empty neighborhoods give undefined rates; all
reported per-community rates are finite, so sampled eligible
neighborhoods are nonempty in every run.

Accuracy, Brier loss, and log loss average over all test records.
Probabilities are clipped to $[10^{-7},1-10^{-7}]$ for these predictive
measures; accuracy uses threshold $.5$ and log loss uses natural
logarithms. For predictions $p_i$ and binary outcomes $y_i$, Brier loss is
$N^{-1}\sum_i(p_i-y_i)^2$, and log loss is
$-N^{-1}\sum_i[y_i\log p_i+(1-y_i)\log(1-p_i)]$.
Table entries are means and standard deviations across the five data
splits and training runs, with divisor five for standard deviations,
conditional on the fixed numerical specifications. Exact eligible-pair
counts and per-community means across seeds are not
reported in these summaries.

\paragraph{Data source.}
The COMPAS data are distributed by ProPublica under its
\href{https://www.propublica.org/datastore/terms/}{Standard Terms of Use},
which require attribution and restrict redistribution. The accompanying
materials do not redistribute the dataset.

\end{document}